\documentclass{article}

\usepackage[preprint]{neurips_2026}

\usepackage{multicol}
\usepackage{subcaption}
\usepackage{comment}
\usepackage[utf8]{inputenc}
\usepackage[T1]{fontenc}
\usepackage{hyperref}
\usepackage{url}
\usepackage{booktabs}
\usepackage{amsmath,amssymb,amsfonts,amsthm}
\usepackage{mathtools}
\usepackage{microtype}
\usepackage{xcolor}
\usepackage{graphicx}
\usepackage{multirow}
\usepackage{quiver}
\usepackage{float}
\usepackage{wrapfig}

\newcommand{\KL}{\mathrm{KL}}
\newcommand{\NLL}{\mathrm{NLL}}

\newcommand{\R}{\mathbb{R}}
\newcommand{\vol}{\text{vol}}

\theoremstyle{plain}
\newtheorem{theorem}{Theorem}[section]

\theoremstyle{definition}

\theoremstyle{remark}

\title{Normalizing Flows on Quotient Manifolds via Boundary Quotients}

\author{%
  William Ghanem \\
  The University of Texas at Austin \\
  \texttt{william.ghanem@my.utexas.edu}
  \And
  Benjamin Cai \\
  The University of Texas at Austin \\
  \texttt{bc38399@my.utexas.edu}
}

\begin{document}

\maketitle

\begin{abstract}
We introduce boundary quotients and present a framework for learning densities on manifolds that arise as boundary quotients of simpler domains. We show that this framework can be used to construct normalizing flows on quotient manifolds
$N/G$, where a discrete group $G$ acts on $N$. We instantiate this construction for genus-$g$ surfaces $\Sigma_g$. When $G$ is finite, we show applicability to symmetry aware learning; we demonstrate this on cyclic quotients of the 3-sphere. Experiments on lens spaces show that simple pre-quotient RealNVP models can achieve strong results while being substantially cheaper to evaluate.
\end{abstract}

\section{Introduction}
Normalizing flows provide a family of invertible generative transformations that enable the approximation of complex probability distributions \cite{kobyzev2020normalizing}. Although classical flow architectures are formulated on Euclidean spaces, many scientific and geometric learning problems are naturally supported on non-Euclidean domains. Indeed, Gemici \textit{et al.} motivate angular variables arising in protein folding, robot limbs, and gene expression as examples where density estimation on manifolds is relevant \cite{gemici2016normalizing}. For instance, Rezende \textit{et al.} develop flows on tori and spheres with applications to multi-link robot arms \cite{rezende2020normalizing} and Köhler \textit{et al.} learn flows on products of tori with applications relevant to Boltzmann generation for molecular systems \cite{kohler2021smooth}. At times, we have seen relevant geometric and probabilistic information built into the flow. For example, Bose \textit{et al.} develop normalizing flows on hyperbolic space that respect the underlying Riemannian geometry \cite{bose2020latent}. Additionally, Katsman \textit{et al.} introduce EMF - a method for learning flows on the intrinsic space of a quotient manifold obtained through group action by flowing along an equivariant vector field obtained as the gradient of an equivariant potential \cite{katsman2021equivariant}. The construction of EMF allows for an equivariant application of CNF \cite{chen2018neural}, which integrates over a vector field and, therefore, requires solving a differential equation. Some more recent flows optimize vector fields by 'matching probability paths' as opposed to solving an ODE. This method was introduced by Ben-Hamu \textit{et al.} in \cite{benhamu2022matching}, with Chen and Lipman extending the method to general geometric manifolds (RFM) in \cite{chen2024flow}. Finally, also relevant is, Manifold Free-Form Flows \cite{sorrenson2024learning}. Here, Sorrenson \textit{et al.} avoid vector fields entirely for faster training by embedding a manifold in $\mathbb{R}^n$, using an earlier inexpensive flow architecture without worry of the image of points in $\mathbb{R}^n$, and projecting to the manifold after training steps.

These methods substantially broaden the scope of manifold generative modeling, but typically require an explicit geometric setup. For example, closed form parametrizations appear in \cite{sorrenson2024learning} and CNF/EMF/RFM require tangent space descriptions and explicit Riemannian metrics. Thus, for spaces naturally presented by a quotient of another space, it makes sense to consider flows on the latter space - especially when this space has simple topology and the subspace of glued points has zero measure.

Another related issue is symmetry. Prior work on equivariant flows argues that when the target density is invariant under a group action, this structure should be built into the generative model rather than left to be learned \cite{kohler2020equivariant}. While recent specific applications incorporate particular symmetries, such as those arising in crystalline materials \cite{miller2024flowmm}, the most general symmetry aware flow is EMF. Our construction provides a general framework in the case that symmetries are given by the action of a discrete group.

In this work, we provide a framework for constructing normalizing flows on quotient spaces. We demonstrate that this construction may be used to construct flows on quotients of manifolds by discrete group actions. Furthermore, we demonstrate that, when symmetry is relevant, the construction implements symmetry awareness in the sense of \cite{kohler2020equivariant} in the strongest sense.
\section{Background}
\subsection{Normalizing Flows}
\label{NF}
In our context of flows, we have two probability spaces $(Z,\mu_Z)$ and $(X,\mu_X)$ where $Z$ and $X$ are the same manifold, $\mu_Z$ is the base prior distribution, and $\mu_X$ is the target distribution. The objective is to train a map $F_\theta:Z \rightarrow X$ ($\theta$ being a finite tuple of trainable parameters) such that the pushforward distribution of the prior approximates the target, that is, $F_{\theta *}\mu_Z \approx \mu_X $. The metric we use to achieve this approximation is the $\KL$ divergence between the probability density function (PDF) of the pushforward $F_{\theta *}\mu_Z$ and the PDF of $\mu_X $. By the change of variable formula, if $F$ is a diffeomorphism, $F_{\theta *}\mu_Z$ locally has PDF 
\[
q_{\theta}(x)=p_Z\!\big(F_\theta^{-1}(x)\big)\ \big|\det J_{F_\theta^{-1}}(x)\big|
\]
and so the $\KL$ divergence between $F_{\theta *}\mu_Z$ and $\mu_X$ is given by 
\[
\KL(F_{\theta*}\mu_Z||\mu_X)=\mathbb{E}_{x \sim F_{\theta *}\mu_Z} \left[ \log\frac{q_{\theta}(x)}{p_X(x)} \right]
\]
In practice, we approximate the above via MC sampling. Hence, by minimizing the expectation formula above with respect to the parameters in $\theta$, we may learn to sample from the target $\mu_X$ by sampling first the prior $\mu_Z$ and, afterwards, applying the learned $F_\theta$. When only samples $\{x_i\}_{i=1}^N$ from $\mu_X$ are available, we instead minimize the negative log-likelihood (NLL)
\[
\NLL(q_\theta)=-\frac{1}{N}\sum_{i=1}^N \log q_\theta(x_i)
\]
where $q_{\theta}$ is as defined above.

\subsection{Topological Preliminaries}
\label{TopologicalPreliminaries}
\begin{wrapfigure}{r}{0.37\textwidth}
  \centering
  \includegraphics[width=0.38\textwidth]{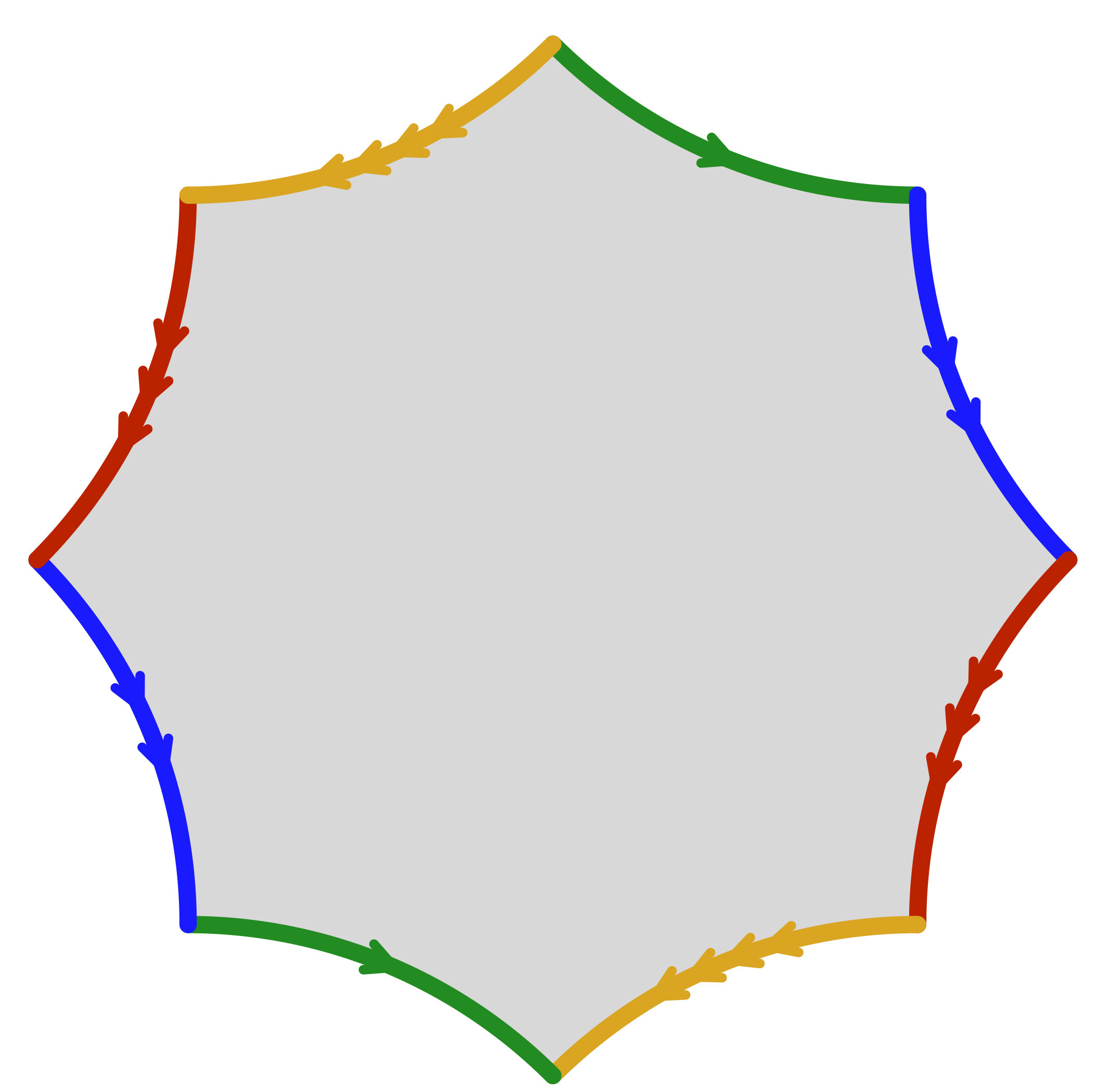}
  \caption{The edge identifications of an $8$-gon in $\mathbb{H}^2$ that makes $\Sigma_2$}
  \label{fig:octagon}
\end{wrapfigure}

We provide the background in topology for this work. Let $M$ be a measurable topological space that is a manifold a.e. (almost everywhere) in the sense that the set of points in $M$ with no Euclidean neighborhood has zero measure in $M$. The most natural example of such a space is a manifold with boundary. We denote this set of points in $M$ with no Euclidean neighborhood by $\partial M$. Let $\sim$ be an equivalence relation on $M$ that identifies only points in $\partial M$. We obtain a new topological space $M/{\sim}$ after quotienting by the equivalence relation. $M/{\sim}$ is a particular instance of a space called a \textit{quotient space}. As we will see, in nice cases, $M/{\sim}$ often becomes a manifold. We call $M/{\sim}$ a \textit{boundary quotient} of $M$ if $M/{\sim}$ is a manifold. In Figure~\ref{fig:octagon}, we provide a canonical case of this phenomenon, where edge identification of an $8$-gon provides the genus-$2$ orientable surface. For more on quotient spaces and quotient topology, we refer the reader to \cite{munkres2000topology}.

A seemingly separate (but as we will see, related) object is a normal covering space. Given a manifold $N$ and a discrete group $G$ acting freely on $N$, we obtain a new space $N/G$ given by identifying orbits in $N$ under the action of $G$. If the action of $G$ is \textit{properly discontinuous} in the sense that for each $x\in N$ there is a neighborhood $U$ of $x$ such that $gU \cap U=\emptyset$ for all nontrivial elements $g$ of $G$, then $N/G$ is a manifold and the projection map, often denoted $\rho :N\rightarrow N/G$, is called a \textit{normal covering space} of $N/G$. The manifold $N/G$ is also an instance of a quotient space. The manifolds $N$ and $N/G$ are exactly the spaces of interest in EMF, though restricted to the case where $G$ is discrete (countable). Assume further that $N$ is a Riemannian manifold and equip $N$ with the induced volume measure. A \textit{fundamental domain} $F\subset N$ is a closed subspace that is a manifold a.e. such that $N= \bigcup_{g\in G}gF$ and $g_1F \cap g_2 F\subset g_1 \partial F\cap g_2\partial F$ whenever $g_1\neq g_2$. We refer to \cite{hatcher2002algebraic} for a complete discussion on normal covering spaces.
\section{Learning equivalence, Boundary Quotients, and Normal Covering Spaces}
In this section, our goal is to provide the mathematical justifications to our approach. In Section~\ref{BoundaryQuotientLearningEquivalence}, we discuss that, at the level of density, it suffices to learn distributions on the manifold $M$ - that is, before identifying points in $\partial M$. In Section~\ref{FromBoundaryQuotientstoNormalCoveringSpaces}, we provide the relationship between boundary quotients and normal covering spaces. In Section~\ref{SymmetryAwarenessLens}, we describe the application to symmetry aware learning.
\subsection{Boundary Quotient Learning Equivalence}
\label{BoundaryQuotientLearningEquivalence}
First, we note that we are in the following setting: $M$ is a measurable topological space that is a manifold a.e. and $M/{\sim}$ is a manifold that is a boundary quotient of $M$. The manifold $M/{\sim}$ is a measurable space with measure given by the pushforward measure through the canonical map $M \rightarrow M/{\sim}$. Given such measurable spaces, we have the vector spaces $L^1(M)$ and $L^1(M/{\sim})$, where $L^1(N)$ is the space of integrable functions defined a.e. on $N$. Two functions $f_1,f_2\in L^1(N)$ are considered equal if $f_1(x)=f_2(x)$ a.e. The term `almost everywhere' is crucial: a statement holds a.e. if the statement is true on a subset $N^{\circ }\subset N$ for which $N\backslash N^{\circ}$ has zero measure. In particular, one often says that $L^1(N)$ consists of equivalence classes of functions, with equivalence classes consisting of functions that are equal a.e. Denote by $P:M\rightarrow M/{\sim}$ the canonical surjective map. Since $M/{\sim}$ is a boundary quotient of $M$ and $\partial M$ has zero measure in $M$, the map $P^{*}:L^1(M/{\sim)}\rightarrow L^1(M)$ given by $f \mapsto f \circ P$ is an isomorphism satisfying $\int_{M/{\sim}}f=\int_{M}P^{*}(f)$. The distributions we wish to approximate are those whose PDFs correspond to functions in $L^1$ with integral one. Such distributions are called $\textit{absolutely continuous}$ distributions. The correspondence $P^{*}$ between $L^1(M/{\sim})$ and $L^1(M)$ gives a correspondence between such distributions. This means that approximating absolutely continuous distributions on $M/{\sim}$ is equivalent to approximating absolutely continuous distributions on $M$. In particular, this is our approach: we approximate distributions on $M/{\sim}$ by considering the distribution as living on $M$.

We note such measure theoretic considerations are not unusual; similar considerations appear either implicitly or explicitly in several manifold flow works \cite{rezende2020normalizing, gemici2016normalizing, sidheekh2022vqflows}. In \cite{kalatzis2021density}, charts may have overlap with positive measure, making a gluing mechanism necessary to the method. In our case, the set of points identified under $\sim$ has zero measure, and so, as per the above discussion, such considerations carry little probabilistic weight. The caveat is giving up global smoothness. The big gain is architectural freedom, providing expressive and inexpensive flows that are smooth a.e.
\subsection{From Boundary Quotients to Normal Covering Spaces}
\label{FromBoundaryQuotientstoNormalCoveringSpaces}
In Section~\ref{BoundaryQuotientLearningEquivalence}, we note that if $M/{\sim}$ is a manifold that is a boundary quotient of $M$, then we may learn an absolutely continuous distribution on $M/{\sim}$ by learning the corresponding distribution on $M$. In this section, we show how this can be used when the manifold of interest is given by the quotient of a discrete group action. We begin with context: we assume that we are given a manifold $N$ and a discrete group $G$ acting properly discontinuously by isometries on $N$. In particular, this means that $\rho:N\rightarrow N/G$ is a normal covering space. We assume that $N$ admits a complete Riemannian metric where \textit{complete} here means that the induced distance metric on $N$ is complete. In this context, we have the following result.
\begin{theorem}
\label{Theorem3.1}
Let $\rho:N\rightarrow N/G$ be a normal covering space between manifolds. Suppose that $N$ admits a complete Riemannian metric and $G$ acts by isometries. Then $\rho$ admits a closed connected fundamental domain $F$. Furthermore, $N/G$ is a boundary quotient of $F$.
\end{theorem}
\begin{proof}
The existence of the fundamental domain $F$ follows directly from Proposition 12 of \cite{kapovich2024properlyDiscontinuousCorrection}. Let $\sim$ denote the equivalence relation on $F$ such that $x \sim y$ if there is $g_1,g_2\in G$ such that $g_1x=g_2y$. Denote by $f:F\rightarrow F/{\sim}$ the canonical surjection. Since $\rho$ is surjective and the orbits of $F$ cover $N$, the restriction map $\rho|_{F}:F\rightarrow N/G$ is surjective. The map $\rho|_{F}$ induces a surjective map $\epsilon:F/{\sim} \rightarrow N/G$ uniquely defined by satisfying the composition $\epsilon \circ f=\rho|_{F}$. Due to the equivalence relation, $\epsilon$ is also injective. Even more is true: by Proposition 34 of \cite{kapovich2024properlyDiscontinuous}, $\epsilon$ is a homeomorphism. We claim that $F/{\sim}$ is a boundary quotient. Indeed, suppose that $x,y\in F$. The two points $x,y$ are identified under $\sim$ if there are $g_1,g_2\in G$ such that $g_1 \neq g_2$ and $g_1x=g_2y$. If this is true, by the definition of fundamental domain, $g_1x \in g_1\partial F$. In particular, multiplying by $g_1^{-1}$, this means that $x\in \partial F$. Similarly, $y\in \partial F$.
\end{proof}
Theorem~\ref{Theorem3.1} combined with the discussion from Section~\ref{BoundaryQuotientLearningEquivalence} determines how one may learn distributions on $N/G$: by identifying the fundamental domain $F$, which necessarily exists. $F$ is a measurable space via restriction of the volume measure on $N$. Using the reasoning in the proof of Theorem~\ref{Theorem3.1}, $N/G$ may be represented by $F/{\sim}$, the boundary quotient of $F$. Following Section~\ref{BoundaryQuotientLearningEquivalence}, approximating distribution on $F$ (or some intermediate quotient of $F$) suffices.

In \cite{kapovich2024properlyDiscontinuousCorrection}, an example of a fundamental domain is given by $\hat D_x=\{y\in N:d(y,x)\leq d(y,gx)\text{ } \forall \text{ } g\in G\}$, where $x\in N$ is any preselected base point. That said, we wish to identify fundamental domains explicitly.
\subsection{Symmetry Awareness}
\label{SymmetryAwareness}
As we mention in the introduction, prior work argues that when symmetry is present, such symmetries should be built into the generative model directly \cite{kohler2020equivariant}. That is, if the manifold $N$ under consideration is acted on smoothly, properly, and freely by a group $G$ via isometries and the target density is invariant under the action (in the sense that $p(gx)=p(x)$ for all $g\in G$), then the flow should exhibit equivariance. In particular, when learning occurs on the ambient manifold $N$ as opposed to the quotient $N/G$, the following quantity should naturally be $\approx 0$:
\begin{align*}
    \mathcal{E}_{sym}=\mathbb{E}_{x \sim q_{\theta}}\big [ \frac{1}{\vol(G)} \int_G \big|\log q_{\theta}(gx)-\log q_{\theta}(x)\big| dg \big ]
\end{align*}
where $q_{\theta}$ denotes the learned density. In EMF \cite{katsman2021equivariant}, learning occurs in the ambient manifold $N$ and equivariance is installed into the CNF architecture using an equivariant potential function and an equivariant prior distribution. Suppose that the group $G$ is discrete and the manifold $N$ is compact (in this case, $G$ must be finite). The discrete version of the term `proper' used above is exactly properly discontinuous. Thus, $N\rightarrow N/G$ is a normal covering space. By Theorem 3.1, there is necessarily a fundamental domain $F$. Given an invariant target distribution on $N$ that we wish to learn equivariantly, the benefit of learning directly on the base manifold $N/G$ using $F$ is as follows: let $p:F\rightarrow \R$ be the learned density on the fundamental domain $F$, then we have an induced density on $N$ given by $p_N(x)=\frac{1}{|G|} \cdot p(g_x x)$ where $g_x\in G$ is such that $g_xx\in F$. In particular, the induced density on $N$ canonically satisfies $\mathcal{E}_{sym}=0$. Therefore, equivariance is installed by picking a suitable learning domain as opposed to building it into the flow architecture, thereby avoiding the architectural constraint equivariance often imposes. In our examples on lens spaces, this allows for the use of simple inexpensive flows in training while still achieving the desired equivariance.
\section{Manifolds as Boundary Quotients}
\label{ManifoldsAsBoundaryQuotients}
In this section, we express two classes of manifolds as boundary quotients: the lens spaces $L(p;q)$ and genus-$g$ surfaces $\Sigma_g$. For both of these classes, each manifold may be expressed as the boundary quotient of a fundamental domain arising from a normal covering space. This makes the examples rich: we may consider target densities/data living initially in the top space (given by $N$ in Section~\ref{TopologicalPreliminaries}) and pushforward to the base space (given by $N/G$ in Section~\ref{TopologicalPreliminaries}) - making the target density intrinsic to $N/G$ and naturally smooth. As we will see, lens spaces are finite cyclic quotients of $S^3$ with fundamental domain parametrized by two functions. Here, we operate precisely in the context of Section~\ref{SymmetryAwareness}, where symmetry awareness is of main interest. In Section~\ref{LensSpacesL(p,q)}, we use EMF \cite{katsman2021equivariant} as a baseline, since its implementation on $S^3$ as an equivariant flow is the most natural choice. In the case of genus-$g$ surfaces, we realize $\Sigma_g$ as the boundary quotient of a $4g$-gon. This $4g$-gon arises as a fundamental domain of a normal covering space $\mathbb{H}^2 \rightarrow \mathbb{H}^2/\Gamma\cong \Sigma_g$, where $\Gamma$ is a Fuchsian group (\ref{FuchsianGroups}). The group $\Gamma$ is infinite discrete here, and so the most relevant sections are \ref{BoundaryQuotientLearningEquivalence} \& \ref{FromBoundaryQuotientstoNormalCoveringSpaces}. To our knowledge, flows on $\Sigma_g$ are yet to be considered in the literature; this construction serves especially in demonstrating generality of the boundary quotient framework beyond lens spaces.
\subsection{Lens Spaces $L(p;q)$}
\label{LensSpacesL(p,q)}
Let $p$ and $q$ be positive integers such that $\gcd(p,q)=1$ and $q < p$. The manifold $L(p;q)$ is obtained as a quotient of $S^3 \subset \mathbb{C}^2$ by the $\mathbb{Z}_p$ action generated by $(z_1,z_2)\mapsto (e^{\frac{2\pi i}{p}}z_1,e^{\frac{2\pi i q}{p} }z_2)$. The action of $\mathbb{Z}_p$ is by isometries. It is also free and properly discontinuous, and so the map $\rho:S^{3}\rightarrow L(p;q)$ is a normal covering space. For $i=1,2$, let $U_i \subset S^3$ be the subspaces defined by $U_i=\{|z_i|^2 \leq \frac{1}{2}:(z_1,z_2)\in S^3\}$, $i=1,2$. Let $r=q^{-1} \text{ mod }p $ and let $s=\frac{rq-1}{p}$. Define maps $f_i$ as follows:
\[
f_1:[0,2\pi] \times D^2 \rightarrow U_1
\]
\[f_1(\theta,\rho e^{i\phi})=\left (\sqrt{\frac{1}{2}}\rho \cdot e^{i(\phi + \frac{r}{p}\theta)},\sqrt{1-\frac{1}{2}\rho^2}\cdot e^{i\frac{\theta}{p}}\right )
\]
and,
\[
f_2:[0,2\pi] \times D^2 \rightarrow U_2
\]
\[f_2(\theta,\rho e^{i\phi})=\left (\sqrt{1-\frac{1}{2}\rho^2}\cdot e^{i\frac{\theta}{p}},\sqrt{\frac{1}{2}}\rho \cdot e^{i(-\phi + \frac{q}{p}\theta)}\right )
\]
Let $F=im(f_1)\cup im(f_2)$, where $im$ denotes the image of a map. The subspace $F$ provides a fundamental domain for the normal covering space $\rho:S^3 \rightarrow L(p;q)$. Let $D_i$ denote the domain of $f_i$, hence each $D_i$ is given by $[0,2\pi]\times D^2$. Since each $f_i$ is a homeomorphism onto its image, we have that $L(p;q)$ is a boundary quotient of $D_1 \cup D_2$. The boundary quotient $F/{\sim}$ from the proof of Theorem 3.1 may be constructed as follows: first, identify $\{0\} \times D^2$ with $\{2\pi\} \times D^2$ in each $D_i$ via the natural identification, resulting in the manifolds $T_i=S^1 \times D^2$ for $i=1,2$. Second, identify $\partial T_1=S^1 \times S^1$ with $\partial T_2=S^1 \times S^1$ along the map $A: \partial T_1 \rightarrow \partial T_2$  given by 
\[\hspace{15pt} \begin{pmatrix} \theta \\ \phi \end{pmatrix}
    \mapsto   \begin{pmatrix}
    r & p \\
    s & q
\end{pmatrix} \begin{pmatrix}
    \theta \\ \phi 
\end{pmatrix}
\mod 2\pi
\]
We use the notation $T_1 \cup _A T_2$ to denote the boundary quotient of $T_1 \cup T_2$ along the map $A$ defined above. As in the proof of Theorem 3.1, $T_1 \cup_AT_2 \cong L(p;q)$ and the homeomorphism is given naturally by descending the maps $f_1$ and $f_2$ to a map from $T_1 \cup_AT_2$ into the orbit space defining $L(p;q)$. In particular, we see that $L(p;q)$ is a boundary quotient of $T_1 \cup T_2$, the disjoint union of two solid tori. For the case of lens spaces, we learn densities on $T_1 \cup T_2$. 
\subsection{Genus-g Surface $\Sigma_g$}
\label{GenusGSurfaceSigmag}
Let $g>1$ be a positive integer and let $n=4g$. We work in the Poincaré disk model for $\mathbb H^2$. Let $P_g\subset \mathbb H^2$ be the regular geodesic $n$-gon centered at the origin with interior angle $2\pi/n=\pi/(2g)$. Equivalently, $P_g$ is the central polygon of the regular $\{4g,4g\}$ tiling of $\mathbb H^2$ (\ref{FuchsianGroups}). Write the vertices of $P_g$ as $v_k=R_g e^{2\pi i k/n}, \text{ }k=0,...,n-1$, where
\[
R_g =
\left(
\frac{\tan(\pi/2-\pi/n)-\tan(\pi/n)}{\tan(\pi/2-\pi/n)+\tan(\pi/n)}
\right)^{1/2}
\]
Let $e_k$ denote the geodesic side joining $v_k$ to $v_{k+1}$, with indices modulo $n$. The genus-$g$ surface $\Sigma_g$ is obtained as a quotient of $\mathbb H^2$ by a torsion free Fuchsian group $\Gamma$. In our construction, $\Gamma$ is
generated by isometries that identify opposite sides of $P_g$. Let
$
\gamma_0,...,\gamma_{2g-1} \in \text{Isom}^+(\mathbb H^2)
$
be orientation preserving hyperbolic isometries such that $\gamma_i$ is the unique isometry of $\text{Isom}^+(\mathbb H^2)$ that sends $e_i$ to $e_{i+2g}$ while sending $v_i$ to $v_{i+2g+1}$ and $v_{i+1}$ to $v_{i+2g}$ (again, indices modulo $n=4g$). Then 
$
\Gamma=\langle \gamma_0,...,\gamma_{2g-1}\rangle
\subset \text{Isom}^+(\mathbb H^2)
$, the subgroup of $\text{Isom}^+(\mathbb{H}^2)$ generated by $\gamma_0,...,\gamma_{2g-1}$. For a more in depth explanation see \ref{FuchsianGroups}. The group $\Gamma$ acts freely and properly discontinuously on $\mathbb H^2$ by isometries. Hence, $\rho:\mathbb H^2\to \mathbb H^2/\Gamma\cong \Sigma_g$ is a normal covering space. As in Section~\ref{LensSpacesL(p,q)}, a fundamental domain may be made explicit: it is exactly the subspace $P_g\subset \mathbb{H}^2$. From here on out, we denote $P_g$ by $F_g$. The boundary quotient $F_g/{\sim}$ from the proof of Theorem~\ref{Theorem3.1} is given by identifying the edge $e_{i}$ of $F_g$ with the edge $e_{i+2g}$, as is shown Figure~\ref{fig:octagon} for the case $g=2$. In particular, the genus-$g$ surface $\Sigma_g$ is a boundary quotient of $F_g$, with identification given by $\sim$. Therefore, for the case of genus-$g$ surfaces, we learn densities on the pre-quotiented
fundamental domain $F_g$.
\section{Experiments}
In this section, we provide experimental results. For each target distribution, the density/data comes from the top space and is pushed forward to the base space. For this, we make use of the explicit maps $N\rightarrow N/G$ and the associated fundamental domain map (see \ref{ManifoldsAsBoundaryQuotients}). As specified in \ref{BoundaryQuotientLearningEquivalence}, we learn on the identified pre-quotient domain, each specified in \ref{ManifoldsAsBoundaryQuotients} and further discussed in \ref{FuchsianGroups}. In Section~\ref{LensSpaceExperiments}, we consider the manifold $L(p;q)$, and in Section~$\ref{GenusGSurfaceExperiments}$, we consider the genus-$g$ surface $\Sigma_g$ 

\subsection{Lens Space Experiments}
\label{LensSpaceExperiments}
We conduct experiments for lens spaces $L(p;q)$. In Section~\ref{SymmetrizedVMF}, we discuss experiments on `symmetrized' vMF distributions. In Section~\ref{SampleTrainingParticle}, we perform a sample-based experiment on particle orientations from the untilted influenza hemagglutinin trimer dataset EMPIAR-10096 of \cite{tan2017addressing}. In Section~\ref{SymmetryAwarenessLens}, we compare with a chart-based $S^3$ flow without symmetry awareness. In the language of Section~\ref{FromBoundaryQuotientstoNormalCoveringSpaces}, $N=S^3$, $G=\mathbb{Z}_p$, and $N/G=L(p;q)$. We use EMF as a baseline, as it is the most natural for a flow on $S^3$ with $\mathbb{Z}_p$ symmetries. We note that our method (PQ RealNVP) uses RealNVP on the pre-quotient space $T_1 \cup T_2$ (see \ref{LensSpaces} for details). Here, PQ stands for `pre-quotient'. 

\subsubsection{Symmetrized vMF Experiments}
\label{SymmetrizedVMF}
\begin{wrapfigure}{r}{0.5\textwidth}
  \centering
  \includegraphics[width=0.5\textwidth]{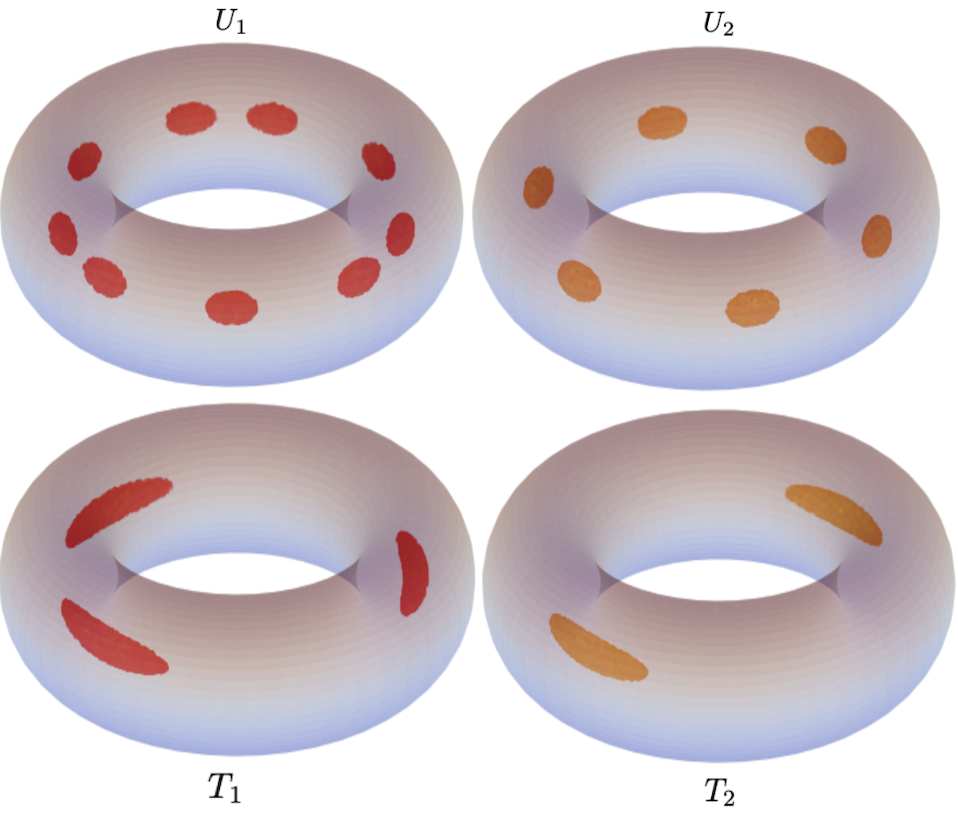}
  \caption{Visualizing symmetry aware learning on $L(3,2)$ using $p^1_{sym}$. The $S^3$ produced $9$ high density modes in $U_1$ and $6$ in $U_2$. PQ RealNVP produces $3$ in $T_1$ and $2$ in $T_2$.}
  \label{fig:symmetryL}
\end{wrapfigure}
The $\text{vMF}(\mu,\kappa)$ PDF on $S^3$ is given by $f(x;\mu,\kappa)=\frac{\kappa}{4\pi^2I_1(\kappa)}e^{\kappa \mu \cdot x}$. This is a density with respect to Riemannian volume measure on $S^3$. By the \textit{symmetrized density} of the PDF $p(x)$ on $S^3$ we mean the density given by $p_{sym}(x)=\frac{1}{p}\sum_{g\in \mathbb{Z}_p}p(gx)$. To achieve the final target density, we take a finite normalized linear combination of vMF densities and then symmetrize. In \ref{KLLensSpace}, Table \ref{tab:lens-space-sweep} shows our findings. Experiment 1 uses $(p,q)=(11,3)$ with target $p^1_{sym}$ where $p^1$ is given in \ref{DensitiesLensSpaces}. For Experiment 2, we use $(p,q)=(7,3)$ with target $p^2_{sym}$ where $p^2$ is given in \ref{DensitiesLensSpaces}. In Experiment 1, the modes are such that they do not intersect the boundary of the learning domain. In Experiment 2, we purposefully place modes on the boundary of the learning domain. We observe that our model favors the former, in terms of both $\KL$ and evaluation time. Regardless of mode placement, however, our model tends to outperform EMF in terms of raw KL. This is a true benefit of our method: we opt for architectural flexibility by building symmetry into the domain construction as opposed to the flow itself. In our experiments, this flexibility results in smaller $\KL$ values in a fraction of the time. The cost, however, is that we disregard artifacts mentioned in \cite{katsman2021equivariant}. In \ref{TopologyLearningL}, we find that continuity of the symmetrized vMF pushforward targets can be captured very close to the boundary identification points, seen visually in Figure \ref{fig:learningtopology}. However, we also find that this may not be done arbitrarily close.
\subsubsection{Sample Based Training: Particle Orientations}

\label{SampleTrainingParticle}
We train our model and EMF on particle orientations from the untilted influenza hemagglutinin trimer dataset EMPIAR-10096 of \cite{tan2017addressing} - a data set consisting of 130,000 axis-angle samples. To do so, we view the data as living on $S^3$ by converting axis angles to unit quaternions (see \ref{Molecularp_foldsymmetries}). The data has a natural $\mathbb{Z}_3$ symmetry about the initial pose; when converting to $S^3$, this corresponds to symmetries given by the group action corresponding to the normal covering space $\rho:S^3 \rightarrow L(6;5)$. We train EMF for 1320 seconds, where it runs through the data once and achieves a final/minimal NLL of $2.26$. We train PQ RealNVP (our model) for 385 seconds. In this time, our model runs through the data set 50 times; it achieves a minimal NLL of $1.98$. In fact, our model achieves an NLL $\leq 2.26$ (the minimal NLL of EMF) after 31 seconds, corresponding to 4 runs through the data. In particular, while we do not claim that EMF is incapable of dropping below our result with several data run-throughs, this result demonstrates that our model is capable of fast and inexpensive symmetry aware learning.
\subsubsection{Symmetry Awareness: Comparison with RealNVP on $S^3$}
\label{SymmetryAwarenessLens}
One of the major benefits of symmetry awareness is that it removes ambiguity caused by density inconsistencies across the same orbit, that is, it allows for $\mathcal{E}_{sym}\approx0$. Furthermore, we observe that when symmetries are well-pronounced in the target distribution, symmetry awareness also makes learning faster and more accurate. We present the example of $p^{3}_{sym}$ where $p^{3}$ is a normalized linear combination of vMF densities defined in \ref{DensitiesLensSpaces}. The density $p^{3}_{sym}$ exhibits $(p,q)=(7,3)$ symmetries. In this case, we see it clearly: consider the Real-NVP style flow given exactly by our flow but learned directly on $S^3=U_1 \cup U_2$ without consideration of actions and fundamental domains. Learning with PQ RealNVP achieves a $\KL$ of $0.09$ after 48 seconds of training. The $S^3$ ablation model still achieves a decent but less impressive final/minimal $\KL$ of $0.82$ after 229 seconds, while having an orbit inconsistency of $\mathcal{E}_{sym}=5.70$. With more symmetry, this disparity in approximation quality becomes even more obvious: the distribution $p^4_{sym}$ exhibits $(p,q)=(100,13)$ symmetries. In this case, PQ RealNVP achieves a $\KL$ of $0.12$ in 20 seconds, while the $S^3$ ablation has $\KL$ hovering around 3.5 past 60 seconds. The $p^3_{sym}$ approximation on $S^3$ also demonstrates the following: a good approximation $\neq$ minimal $\mathcal{E}_{sym}$. In Figure~\ref{fig:symmetryL}, we demonstrate symmetry awareness visually for $(p,q)=(3,2)$ and density $p^1_{sym}$: the $S^3$ ablation gives $9$ modes in $U_1$ and $6$ modes in $U_2$. PQ Real-NVP gives $9/p=3$ modes in $T_1$ and $6/p=2$ modes in $T_2$.
\subsection{Genus-g Surface Experiments}
\label{GenusGSurfaceExperiments}
\begin{wrapfigure}{r}{0.5\textwidth}
  \centering
    \includegraphics[width=0.5\textwidth]{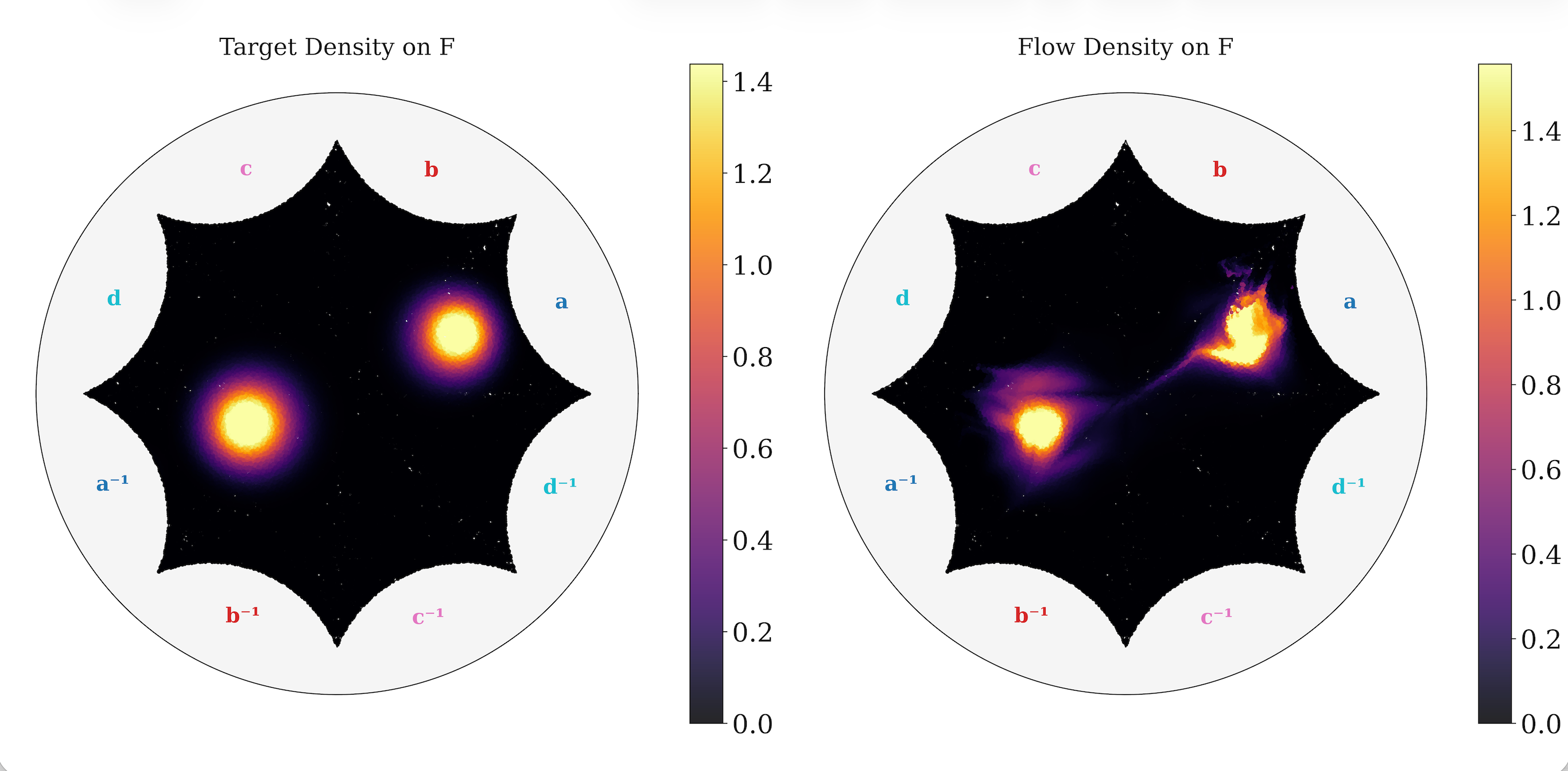}
    \caption{Hyperbolic Gaussian with equally weighted centers at $(0.4, 0.2); (-0.3, -0.1)$ and $\sigma =0.2$ showing multi-modal learning.}
    \label{fig:gaussian_twomodes}
\end{wrapfigure}
In this section, we consider the closed genus-$g$ surfaces $\Sigma_g$, realized as the quotient $\mathbb {H}^2/\Gamma$ of the hyperbolic plane $\mathbb{H}^2$ by the cocompact Fuchsian group $\Gamma$. In the language of Section~\ref{FromBoundaryQuotientstoNormalCoveringSpaces}, $N=\mathbb{H}^2$, $G=\Gamma$, and $N/G=\Sigma_g$. We learn on a fundamental domain given by a $4g$-gon (see \ref{FuchsianGroups}). As a motivating class of target distributions, we consider hyperbolic heat kernel distributions on $\mathbb H^2$ and use the pushforward to $\Sigma_g$ as the target - providing a heat kernel distribution on $\Sigma_g$. Explicit formulas for heat kernels on hyperbolic space are given in \cite{grigoryan1998heat}, and the use of heat kernels on $\mathbb{H}^2$ to generate heat kernels on $\Sigma_g$ is discussed further in \cite{jones2010heat}. In Section~\ref{HyperbolicHeatKernel}, we approximate the induced heat kernel distributions. In Section~\ref{HeatSemigroupIdentity}, we use the learned distribution to numerically test the heat semigroup identity $(e^{t\Delta_{\Sigma_g}}f)(x)= \mathbb E_x[f(B_t)]$, where $B_t$ denotes Brownian motion on $\Sigma_g$ generated by the Laplace-Beltrami operator \cite{grigoryan2002hitting}. We do this using approximations in \ref{HyperbolicHeatKernel}, since $B_t$ has density the pushforward hyperbolic heat kernel density on $\Sigma_g$ \cite{grigoryan1998heat,jones2008heat}. Figure \ref{fig:gaussian_twomodes} shows the flow learning a multimodal distribution, and the construction of the Hyperbolic Gaussian is explicitly explained in \ref{HyperbolicGaussianConstruction}.

\subsubsection{Hyperbolic Heat Kernel} 
\label{HyperbolicHeatKernel}
The Hyperbolic Heat Kernel $K_t(x,y)$ defined on $\mathbb{H}^2$ gives the heat at $y$ after time $t$ from a unit source of heat located at $x$. If we set $r = d_\mathbb{H}(x,y)$, we may write $K_t(x,y)$ in terms of $r$, resulting in a function $p_t(r)$ such that $p_t(r)=K_t(x,y)$ \cite{grigoryan1998heat}. The function $p_t(r)$ is a density on $\mathbb{H}^2$ with respect to the hyperbolic volume measure, and is given by
\begin{align*}
    & p_t(r) = \frac{\sqrt{2}}{(4\pi t)^{3/2}} e^{-t/4}\int_r^\infty \frac{s \, e^{-s^2 / 4t}}{\sqrt{\cosh s - \cosh r}} \, ds
\end{align*}
By periodization over a Fuchsian group $\Gamma$, we obtain the heat kernel on $\Sigma_g$ \cite{jones2010heat}  
\begin{align*}
    & K_t^{\Sigma}(x,y)=p_t^{\Sigma_g}(x,y) = \sum_{\gamma \in \Gamma} p_t(d_{\mathbb{H}}(x,\gamma y))
\end{align*}

\begin{wrapfigure}{r}{0.5\textwidth}
  \centering
    \includegraphics[width=0.5\textwidth]{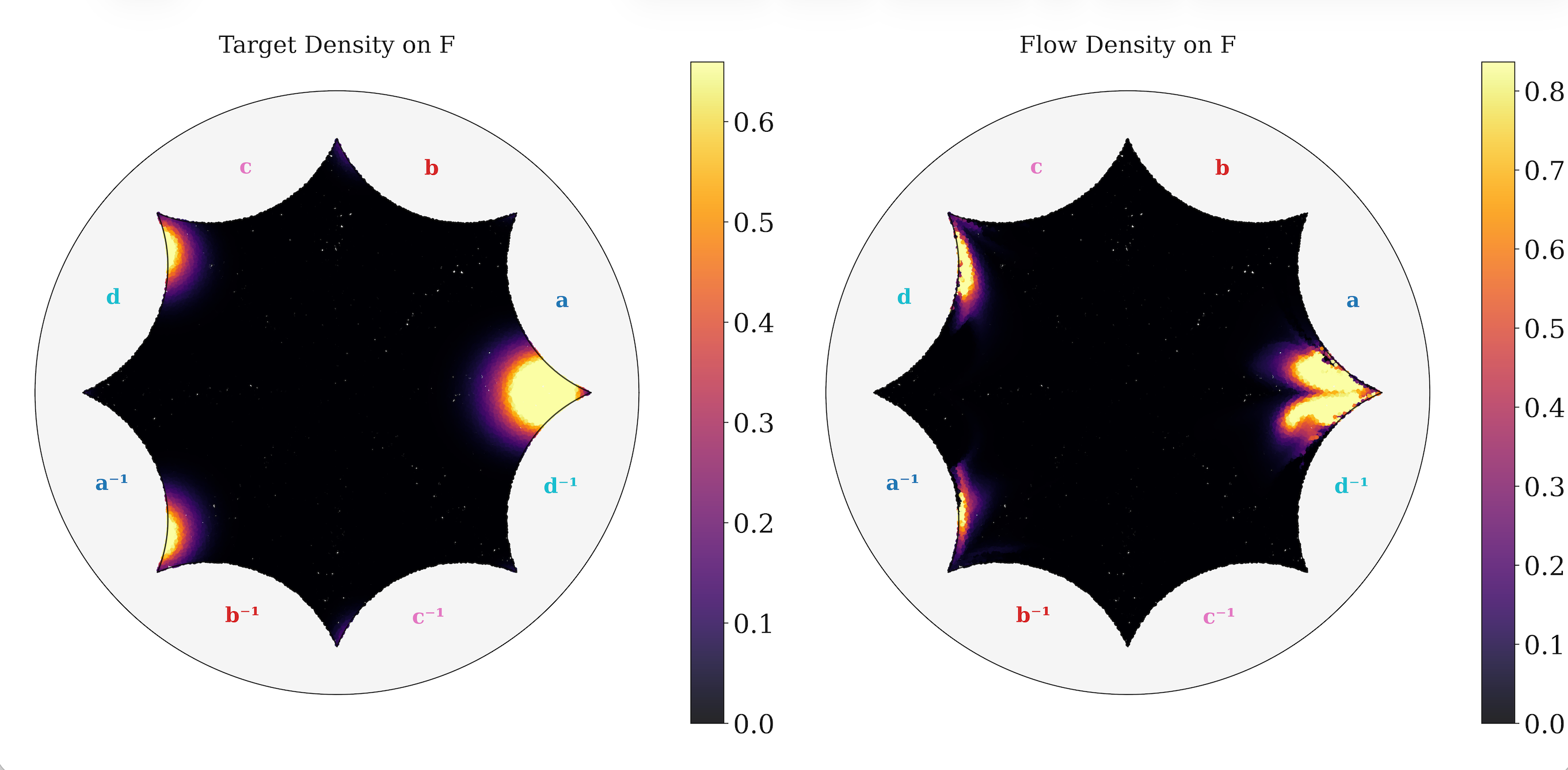}
    \caption{Hyperbolic heat kernel with $t=0.1, x = (0.7,0.0)$ on $\Sigma_2$ or the $\{8,8\}$ fundamental polygon showing density crossing edges.}
    \label{fig:heat_kernel_t01_c0700}
\end{wrapfigure}

The function $p_t^{\Sigma_g}(x,y)$ is a density in the $y$ variable (we keep $t,x$ fixed) and is exactly the pushforward density on $\Sigma_g$ from $p_t(r)$ living naturally on $F_{g}$. We approximate using a RealNVP style flow (\ref{GenusGSurface}). In \ref{KLTableGenusG}, Table~\ref{tab:KLcomparisonheatkernel} shows $\KL$ values for different $t,x$. We achieve accurate approximations, even when mass is located near the boundary. That said, learning is favored when there is less mass near the boundary of $F_g$. In Figure~\ref{fig:heat_kernel_t01_c0700}, we include a visualization with $t,x=0.1,(0.7,0.0)$. As referenced in \ref{TopologyLearningL}, Figure~\ref{fig:heat_kernel_t01_c0700} shows that the correct topology is learned near the boundary, with high mass regions near the boundary being reflected on both sides of the designated boundary identification. The same issues arise sufficiently close to the boundary, however. We also observe artifacts created by non-differentiabilities of the map $F_{g}\rightarrow D^2$ used in the flow construction on the rays from the origin to the vertices of the $4g$-gon. Finally, we observe that the choice of $g$ has little effect on approximation quality. Figure~\ref{fig:heatkernel16gon} in \ref{VisualizationSigma4} showcases learning on a $g > 2$ surface.

\subsubsection{Heat Semigroup Identity}
\label{HeatSemigroupIdentity}
The heat semigroup identity gives that 
\[
(e^{t\Delta_{\Sigma_g}}f)(x)=\mathbb E_x[f(B_t)]=\int_{\Sigma_g}f(y)p_t^{\Sigma_g}(x,y)d\text{vol}_{\Sigma_g}(y)
\]
That is, if $B_{t,x}\sim p_t^{\Sigma_g}(x,y)$, then $(e^{t\Delta_{\Sigma_g}}f)(x)=\mathbb E[f(B_{t,x})]$. Using our learned sampling, we may approximate 
\[
(e^{t\Delta_{\Sigma_g}}f)(x)\approx \frac{1}{N} \sum_{i=1}^N f(b_i)
\] 
where each $b_i$ is sampled from $B_{t,x}$. It is known that if $f(y)=K_s^{\Sigma_g}(a,y)$ for fixed $s,a$, then $(e^{t\Delta_{\Sigma_g}}K_s^{\Sigma_g}(a,\cdot))(x)=K_{s+t}^{\Sigma_g}(a,x)$ \cite{grigoryan2009heat}. Hence, if we let $s=t$, we have $(e^{t\Delta_{\Sigma_g}}K_t^{\Sigma_g}(a,\cdot))(x)=K_{2t}^{\Sigma_g}(a,x)$. Therefore, for this choice of $f$, the heat semigroup identity is equivalent to $\mathbb E[K_t^{\Sigma_g}(x_0,B_{t,x})]=K_{2t}^{\Sigma_g}(x_0,x)$. In \ref{RelErrorGenusG}, we include Table \ref{tab:HeatSemigroup} where we verify this final formula using our approximation of $B_{t,x}$. We verify for varying values of $t,x,x_0$. We compute relative error between the approximation using sampling and the true value. We observe that the best approximations typically occur when $x$ is at the origin. When $x\in F_{g}$ is far from the origin, the best results are given when $x_0$ is closer to $x$. This is expected: for instance, Table~\ref{tab:KLcomparisonheatkernel} shows that for $t=0.05$, $x=(0.5,0.5)$, and $g=2$, approximation yields a final $\KL$ of $0.64$. In particular, certain regions may be difficult to approximate, especially those with less probabilistic importance (i.e. points away from $x$). This is reflected by Table~\ref{tab:HeatSemigroup}, where relative error increases for points $x_0$ further from $x$. However, most relative error values are less than 0.1, showing our implementation indeed approximates the heat semigroup property. Figure~\ref{fig:heat_kernel_relErr} presents heatmaps of relative error for the heat semigroup property of select hyperbolic heat kernel distributions.

\section{Limitations}
\label{Limitations}
Our contribution trades global smoothness for architectural flexibility: since we learn on the pre-quotient domain, the learned density need not exhibit global smoothness. In practice, this may produce local artifacts near boundary regions not otherwise present in the true target distribution, as uncovered in various figures and \ref{TopologyLearningL}. Our genus-$g$ implementation also relies on a particular parametrization of the fundamental domain $F_g$, which introduces non-differentiabilities along rays to the vertices. Finally, while Theorem~\ref{Theorem3.1} provides a fundamental domain for any normal covering space, the practical implementation of our method requires identifying this domain explicitly, and therefore depends on detailed knowledge of the covering space under consideration.

\section{Conclusion}
\label{Conclusion}
We introduce a framework for learning distributions on manifolds that arise as boundary quotients. We demonstrate that this method may be applied to manifolds arising as quotients by discrete group actions. This gives the central progression of the paper: boundary quotients provide the principle, normal covering spaces provide the source of examples, and in the compact finite group setting we may obtain symmetry awareness by choosing a learning domain rather than constraining the architecture. On lens spaces, we demonstrate that this approach allows simple RealNVP style flows to achieve accurate and efficient symmetry aware learning. On genus-$g$ surfaces, we show that this framework can approximate heat kernel distributions and numerically recover the heat semigroup identity. Together, these results suggest that quotient topology can serve as a useful design principle for generative modeling on non-Euclidean spaces.

\bibliographystyle{plainnat}
\bibliography{references}

\newpage
\appendix

\section{Additional Details}
\subsection{Constructing the Target Distribution from $N\rightarrow N/G$}
\label{ConstructingTargetFromN}
Let $\rho:N \rightarrow N/G$ be a normal covering space satisfying the conditions of Theorem~\ref{Theorem3.1}. Let $F$ be a fundamental domain for $\rho$. Suppose we have a distribution $\mu_N$ on $N$ and that we wish to approximate $\rho_*\mu_N$ using flows. As explained in \ref{BoundaryQuotientLearningEquivalence}, we wish to use $F$ as a learning domain. Let $q:F \rightarrow N/G$ be the map $\rho|_{F}$. Recall from \ref{BoundaryQuotientLearningEquivalence} that we have a correspondence $\{\text{absolutely continuous distributions on $F$}\} \leftrightarrow \{\text{absolutely continuous distributions on $N/G$}\}$. We wish to identify and learn the unique distribution $\mu_F$ on $F$ such that $q_*\mu_F=\rho_*\mu_N$. In the case that the density $p_N$ of $\mu_N$ is known, $\mu_F$ has density with respect to the restricted volume measure given by $p_F(x)=\sum_{g\in G}p_{N}(gx)$. Remark that when $N$ is compact, $G$ is finite, and $p_N$ is symmetric with respect to the action of $G$ (that is, $p_N(gx)=p_N(x)$ for all $g\in G$), we have that $p_F(x)=|G|\cdot p_N(x)$. In the case of $L(p;q)$, we utilize this formula. When samples $\{x_i\}$ on $N$ with respect to $\mu_N$ are given, we construct a data set with respect to $\mu_F$ given by $\{g_ix_i\}$ where $g_i\in G$ is an element of $G$ for which $g_ix_i\in F$. In either case, we train on $F$ using the altered density/samples. Finally, we note that when $G$ is infinite, we cannot sum over infinitely many elements. In this case, we sum over a finite number of elements until target density on $F$ has integral $\approx 1$. In the case of $\mathbb{H}^2 \rightarrow \Sigma_g$, we choose a fixed $l>1$ and perform the sum in $p_{F}$ over all combinations of generators $\gamma_{i_1}^{\pm1}\cdot ...\cdot \gamma_{i_m}^{\pm1}$ with $m\leq l$. We avoid potential repeats by keeping track of where the origin is sent for each used combination.

\subsection{The Fuchsian Group $\Gamma$}
\label{FuchsianGroups}
In Section~\ref{LensSpacesL(p,q)}, we gave an explicit description of the $\mathbb{Z}_p$ action on $S^3$ for the normal covering space $S^3 \rightarrow L(p;q)$. We now provide an explicit description of the generators $\gamma_0,...,\gamma_{2g-1}$ of the Fuchsian group $\Gamma$. We follow the construction from \cite{stepanyants2025diameter}. Let $n=4g$ and let $v_k=R_g e^{2\pi i k/n}, \text{ } k=0,...,n-1$ be the vertices of the regular hyperbolic $n$-gon $F_g=P_g\subset \mathbb H^2$, with indices taken modulo $n$. Let $e_i$ denote the geodesic side joining $v_i$ to $v_{i+1}$. For $i=0,...,2g-1$, we pair $e_i$ with the opposite side $e_{i+2g}$ by an orientation preserving hyperbolic isometry $\gamma_i \in \text{Isom}^+(\mathbb H^2)$ such that $\gamma_i(v_i)=v_{i+2g+1}, \text{ } \gamma_i(v_{i+1})=v_{i+2g}$. We write $\gamma_i$ explicitly in the Poincaré disk model of $\mathbb{H}^2$. Let
$z_i^-=v_i,\text{ } z_i^+=v_{i+1},\text{ }
w_i^-=v_{i+2g+1},\text{ and } w_i^+=v_{i+2g}.
$
Let $A_i$ and $B_i$ be disk automorphisms given by
\[
A_i(z)=\frac{z-z_i^-}{1-\overline{z_i^-}z},
\qquad
B_i(z)=\frac{z-w_i^-}{1-\overline{w_i^-}z}.
\]
Remark that $A_i(z_i^-)=0$ and $B_i(w_i^-)=0$. Let
\[
\lambda_i=\frac{B_i(w_i^+)}{A_i(z_i^+)}
\]
Since the paired sides have the same hyperbolic length and $A_i(z_i^-)=B_i(w_i^-)=0$, we have that $|\lambda_i|=1$. The edge pairing isometry $\gamma_i$ is given by
\[
\gamma_i(z)=B_i^{-1}\!\left(\lambda_i A_i(z)\right)
\]
where
\[
B_i^{-1}(z)=\frac{z+w_i^-}{1+\overline{w_i^-}z}.
\]
is the inverse of $B_i(z)$. Note that $\gamma_i$ is indeed an isometry of the disk model of $\mathbb{H}^2$, since $A_i$ and $B_i$ are both isometries and $|\lambda_i|=1$.
This isometry satisfies $\gamma_i(z_i^-)=w_i^-$ and $\gamma_i(z_i^+)=w_i^+$. That is, $\gamma_i$ satisfies 
$\gamma_i(v_i)=v_{i+2g+1}$ and $\gamma_i(v_{i+1})=v_{i+2g}$. The inverse edge pairing map is 
\[
\gamma_i^{-1}(z)=A_i^{-1}\left(\lambda_i^{-1}B_i(z)\right)
\]
where
\[
A_i^{-1}(z)=\frac{z+z_i^-}{1+\overline{z_i^-}z}.
\]
is the inverse of $A_i(z)$.

\begin{wrapfigure}{r}{0.45\textwidth}
   \centering
    \includegraphics[width=0.467\textwidth]{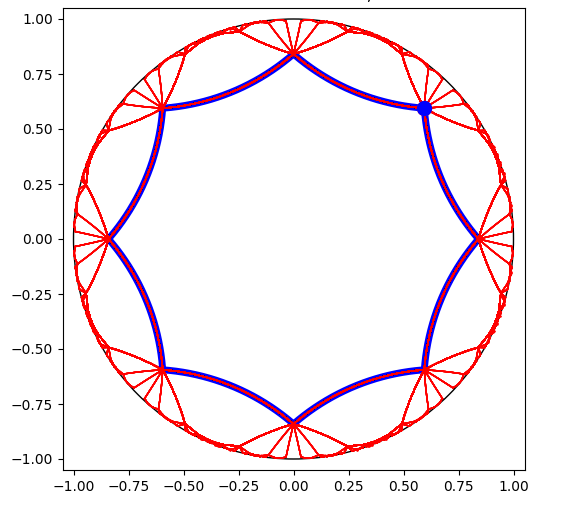}
    \caption{The $\{8, 8\}$ tiling of $\mathbb{H}^2$}
    \label{fig:tiles}
\end{wrapfigure} 

Geodesic segments between points may be given by a closed formula, but we use the orthogonal circle construction \cite{anderson2005hyperbolic}.

For tilings of $\mathbb{H}^2$ using regular polygons referred to in Section~\ref{GenusGSurfaceSigmag}, it is standard to refer to $\{p, q\}$ tilings where $p$ is the number of edges of each polygon and $q$ is the number of edges emanating from each vertex \cite{frettloh2008counting}. In Figure \ref{fig:tiles}, we present the $\{8, 8\}$, which corresponds to the genus-$2$ case. In the $\{4g,4g\}$ tiling, each generator $\gamma_i$ of the Fuchsian group together with its inverse sends $F_g$ to one of the $4g$ adjacent 4g-gons. For example, for $g=2$, the blue central $8$-gon polygon in Figure~\ref{fig:tiles} is mapped to one of the $8$ side-adjacent red $8$-gons under each $\gamma_i^{\pm 1}$, $i=0,...,3$.

\subsection{Pre-Quotient Flow Construction}
\label{PreQuotientFlow}
In this section, we demonstrate how flows are constructed on the domains in which $L(p;q)$ and $\Sigma_g$ arise as a boundary quotient.
\subsubsection{Lens Space $L(p;q)$}
\label{LensSpaces}
Recall from Section~\ref{LensSpacesL(p,q)} that we learn on $T_1 \cup T_2$, two copies of $S^1 \times D^2$. We learn using a flow on each $T_i$ separately. As the prior distribution on $T_i$, we use $\mu_{Z_i} \sim VM_{[0,2\pi)}(0,\kappa_i)\times N(0,\sigma_i^2 I_2)|_{D^2}$. For the $S^1$ product, we use a von Mises distribution in $[0,2\pi)$ centered at $0$ with concentration $\kappa_i$ and in the $D^2$ product, we use a truncated normal distribution centered at the origin with variance $\sigma_i^2$. Its PDF is given by
\[
p_{Z_i}(\theta,x,y)=\frac{e^{\kappa_i \cos(\theta)}}{2\pi I_0(\kappa_i)} \cdot \frac{e^{-\frac{x^2+y^2}{2\sigma_i^2}}}{2\pi \sigma_i^2(1-e^{-1/(2\sigma_i^2)})}
\]
for $(\theta,x,y)\in [0,2\pi) \times D^2$. We use a RealNVP style flow \cite{dinh2017density}. Learnable maps are coupling layers $[0,2\pi)\times\mathbb{R}^2\rightarrow [0,2\pi)\times \mathbb{R}^2$, of which there are two kinds. Formally, the layer keeping the coordinate corresponding to $S^1$ fixed is
\[
c_1(y_1,y_2)=\left (y_1,y_2\odot e^{s_1(\cos(y_1),\sin(y_1))}+t_1(\cos(y_1),\sin(y_1)) \right )
\]
and, keeping the coordinates corresponding to $D^2$ fixed
\[
c_2(y_1,y_2)=(y_1+t_2(y_2)\text{ mod }2\pi,y_2)
\]
Above, the $s_1$ and $t_i$ are trainable linear neural networks of appropriate size. Specifically:
\[
 s_1: \text{Linear}(2,\text{hdim}=32) \rightarrow \text{ReLU}   \rightarrow \text{Linear}(\text{hdim}=32,2) \rightarrow \text{TanH} 
\]
\[
 t_1: \text{Linear}(2,\text{hdim}=32) \rightarrow \text{ReLU}  \rightarrow \text{Linear}(\text{hdim}=32,2) 
\]
\[
 t_2: \text{Linear}(2,\text{hdim}=32) \rightarrow \text{ReLU}  \rightarrow \text{Linear}(\text{hdim}=32,1)\rightarrow \text{TanH} 
\]
Let $\psi:\mathbb{R}^2\rightarrow int(D^2)$ be the diffeomorphism defined by $\psi(v)=\frac{v}{1+||v||}$. The forward diffeomorphism $F_{\theta}:S^1\times D^2 \rightarrow S^1\times D^2$ is therefore defined as:
\[
(\theta,v)  \mapsto (\theta,\psi^{-1}(v)) \xrightarrow{c_1^1c_2^1...c_1^nc_2^n}  (\theta_c,v_c) \mapsto (\theta_c,\psi(v_c))
\]
For training when a target density is given, we construct the target density $q$ on $T_1 \cup T_2$ with respect to the Lebesgue measure using the maps defined in Section~\ref{LensSpacesL(p,q)} and the general method in \ref{ConstructingTargetFromN}. Since we train on each $T_i$ independently, we define densities $q_i$ on $T_i$ such that $q=(1-I_2)q_1+I_2q_2$ where $I_2=\int_{T_2} qd\theta dx dy$. After training, we recouple both learned distributions on each $T_i$ using the same constant $I_2$ to obtain an approximation of $q$. That is, on each $T_i$, the final approximation is given by $(1-I_2)F_{\theta*}Z_1+I_2F_{\theta_*}Z_2$ where $Z_i$ denotes the prior on $T_i$. On each $T_i$, we use the loss function the $\KL$ divergence defined in Section~\ref{NF} as the loss function.

Sample based training uses the same concept: we move the data to $T_1 \cup T_2$, define $I_2 =\frac{\# x_i \text{ in }T_2}{\# x_i}$ and the approximation is given by $(1-I_2)F_{\theta*}Z_1+I_2F_{\theta_*}Z_2$. For sample based training, the loss function on each $T_i$ uses $\NLL$, as defined in Section~\ref{NF}.

For density based training, we use $\sigma_i=0.25$ and $\kappa_i=5$. In sample based training, we approximate a uniform distribution via $\sigma_i=10$ and $\kappa_i=0.1$. Throughout, we train with Adam using learning rate $10^{-3}$, batch size
$1000$, and 18 alternating coupling layers on a T4 GPU with 16 GB VRAM.

\subsubsection{Genus-g Surface $\Sigma_g$}
\label{GenusGSurface}
In implementation, we parameterize $F_g$ by the unit disk using the map $\Psi_g:F_g\to D^2$ defined by

\[
\Psi_g(\rho e^{i\theta})=\frac{\rho}{r_g(\theta)}e^{i\theta}, \qquad \text{det} (D \Psi_g) = \frac{1}{r_g(\theta) ^ 2}
\]

where $r_g(\theta)$ is defined as the Euclidean distance from the origin to unique intersection point of $\{te^{i\theta}:t\geq 0\}$ and $\partial F_g$. This gives inverse: 

\[
\Psi_g^{-1}(r e^{i\theta})= r \cdot  {r_g(\theta)}e^{i\theta}, \qquad \text{det} (D \Psi_g^{-1}) = r_g(\theta) ^ 2
\]

Hence learning on $F_g$ is implemented as learning on $D^2$, together with the corresponding change of variables factor. 

We describe learning on $D^2$, where we implement a RealNVP style flow \cite{dinh2017density}. We use a sequence of coupling layers; in this case, as functions $\mathbb{R}^2 \rightarrow \mathbb{R}^2$. Explicitly, we use maps

\[ c_1(y_1, y_2) = (y_1, y_2 \odot \exp(s_1(y_1)) + t_1(y_1)) \]
\[c_2(y_1, y_2) = (y_1 \odot \exp(s_1(y_2)) + t_1(y_2), y_2)\]

where $s_1, t_1$ are linear neural networks. Specifically:

\[s_1 : \text{Linear}(1,\text{hdim}=32) \rightarrow \text{ReLU}  \rightarrow \text{Linear}(\text{hdim}=32,1) \rightarrow \text{TanH} \]
\[t_i : \text{Linear}(1,\text{hdim}=32) \rightarrow \text{ReLU}  \rightarrow \text{Linear}(\text{hdim}=32,1)\]

Let $\psi: \mathbb{R}^2\rightarrow int(D^2)$ be the diffeomorphism $\psi(v) = \frac{v}{1 + ||v||}$. The forward diffeomorphism $F_{\theta}$ is therefore defined as:
 
\[
(x,y)  \mapsto \psi^{-1}\left( (x,y) \right)\xrightarrow{c_1^1c_2^1...c_1^nc_2^n}  (x_c,y_c) \mapsto \psi\left( (x_c,y_c)\right)
\]

The loss function used is then $\KL$ divergence Section~\ref{NF} between the pushforward prior and the target pushforward density from \ref{ConstructingTargetFromN} moved to $D^2$ via $\Psi_g$.

We use Adam optimizer and train on a T4 GPU with 16 GB of VRAM. The model was trained for 1000 steps with a batch size of 1000 and a learning rate of 0.001. We use a depth of 18 alternating coupling layers.

For the base distribution, we use a Bernoulli mixture of a Beta distribution and a truncated Gaussian distribution. The Gaussian density is given by

\[
p_{N}(z; \sigma) = \frac{1}{Z_{\sigma}} \exp\!\left(-\frac{\|z\|^2} {2\sigma^2}\right),
\qquad
Z_{\sigma} = 2\pi\sigma^2(1-e^{-1/(2\sigma^2)})
\]

The Beta density is given by

\[
p_{B}(z; a, b) = \frac{1}{2\pi B(a,b)} \, \|z\|^{a-2}(1-\|z\|)^{b-1},
\qquad 
B(a,b) = \frac{\Gamma(a)\Gamma(b)}{\Gamma(a+b)}
\]

The Bernoulli mixture is therefore given by

\[
p(z; c, \sigma, a, b) = c \, p_{{N}}(z; \sigma) + (1-c)\, p_{{B}}(z; a, b)
\]

For training, we use $c = 0.6, \sigma = 0.3, a = 4, b = 1$.

\subsection{Equivariant Manifold Flow Baseline}
\label{EMFBaseline}
We implement an Equivariant Manifold Flow (EMF) baseline following
\cite{katsman2021equivariant} for comparison with our $L(p;q)$ flow. The model is defined on $S^3\subset \mathbb{R}^4$ with cyclic $\mathbb{Z}_p$ action 
\[
g_k\cdot (z_1,z_2)
=
\left(e^{2\pi i k/p}z_1,\ e^{2\pi i qk/p}z_2\right),
\text{ } k=0,\ldots,p-1
\]
To obtain an invariant potential, we average a scalar neural network over the entire finite group action
\[
\Phi_\theta(x,t)=\frac{1}{p}\sum_{k=0}^{p-1}\phi_\theta(g_k\cdot x,t)
\]
The vector field on $S^3$ is the tangent projection of the Euclidean gradient of this potential \cite{AbsMahSep2008}:
\[
X_\theta(x,t)
=
\nabla_x \Phi_\theta(x,t)
-
\langle \nabla_x\Phi_\theta(x,t),x\rangle x
\]
The construction requires an equivariant prior distribution - we use the uniform distribution on $S^3$ as the base distribution. We train the resulting manifold CNF using the change-in-density equation \cite{mathieu2020riemannian}
\[
\frac{d}{dt}\log q_t(x_t)
=
-\text{div}_{S^3}X_\theta(x_t,t)
\]
When the target density $\pi$ is available, we
train by minimizing the $\KL$ divergence at time $t=1$
\[
\KL=\frac{1}{N}\sum_{i=1}^N
\left(\log q_\theta(x_1^{(i)})-\log \pi(x_1^{(i)})\right),
\]

where $q_{\theta}$ denotes the solution to change-in-density equation at $t=1$ and samples obtained by flowing the uniform base distribution samples. For
sample based learning, we integrate the flow backward from data samples and minimize negative log-likelihood.

In all experiments, $\phi_\theta$ is an MLP with TanH activations
and hidden dimensions $64$ of the form

\[
\text{Linear}(5,64) \rightarrow \text{TanH}  \rightarrow \text{Linear}(64,64) \rightarrow \text{TanH}\rightarrow \text{Linear}(\text{hdim}=64,1)
\]

We train with Adam using learning rate $10^{-3}$, batch size
$256$, and RK4 integration with step size $0.05$ on the interval $[0,1]$ on a T4 GPU with 16 GB VRAM.

\subsection{Multi-modal distribution in Figure \ref{fig:gaussian_twomodes}}
\label{HyperbolicGaussianConstruction}
We define the multi-modal distribution used in Figure~\ref{fig:gaussian_twomodes}. We define a weighted sum of Gaussians with centers $c_1, c_2,...,c_K$ and weights $w_1, w_2, ..., w_K$ where $\sum_{k = 1}^K w_k = 1$ given by:

\[
p_{\Sigma_g} (z) = \sum_{\gamma \in \Gamma} \sum_{k=1}^K w_k \exp \left ( - \frac{d_{\mathbb{H}^2} (\gamma \cdot z,c_k)^2}{2\sigma ^ 2} \right)
\]

\subsection{The densities $p^i$ on $S^3$}
\label{DensitiesLensSpaces}
We define the densities $p^i$ on $S^3$ from 5.1.1/5.1.3. 

The density $p^1$ is given by:
\[
p^{1}(x)=\frac{1}{2}\big (\frac{1}{2}(f(x;c_1,35)+f(x;c_2,35)) + \frac{1}{3}(f(x;b_1,35)+ f(x;b_2,35)+f(x;b_3,35))\big )
\]
\[
c_1=(1,0,0,0); \quad c_2=(\cos(\pi/3), \sin(\pi/3),0,0)
\]
\[
b_1=(0,0,1,0); \quad b_2=(0,0,\cos(\pi/4), \sin(\pi/4)); \quad b_3=(0,0,\cos(\pi/2),\sin(\pi/2)) 
\]
The density $p^2$:
\[
 p^{2}(x)=\frac{1}{3}\big (\frac{1}{3}(f(x;c_1,65)+\frac{2}{3}f(x;c_2,55)\big ) + \frac{2}{3}\big (\frac{2}{3}(f(x;b_1,65)+ \frac{1}{3}f(x;b_2,80))\big )
\]
\[
c_1=(\sqrt{\frac{1}{2}},0,\sqrt{\frac{1}{2}},0); \quad c_2=(\sqrt{\frac{1}{2}}\cos(\pi/7),\sqrt{\frac{1}{2}}\sin(\pi/7),\sqrt{\frac{1}{2}},0)
\]
\[
b_1=(0,\sqrt{\frac{1}{2}},\sqrt{\frac{1}{2}},0); \quad b_2=(\sqrt{\frac{1}{4}},\sqrt\frac{1}{4},\sqrt{\frac{1}{2}}\cos(4\pi/21), \sqrt{\frac{1}{2}}\sin(4\pi/21)) 
\]
The density $p^3$ is given by the same definition as $p^2$ but with:
\[
c_1=(\sqrt{\frac{2}{3}},0,\sqrt{\frac{1}{3}},0); \quad c_2=(\sqrt{\frac{3}{4}}\cos(\pi/7),\sqrt{\frac{3}{4}}\sin(\pi/7),\sqrt{\frac{1}{4}},0)
\]
\[
b_1=(0,\sqrt{\frac{1}{4}},\sqrt{\frac{3}{4}},0); \quad b_2=(\sqrt{\frac{1}{7}},\sqrt\frac{1}{7},\sqrt{\frac{5}{7}}\cos(4\pi/21), \sqrt{\frac{5}{7}}\sin(4\pi/21)) 
\]
The density $p^4$:
\[
 p^{4}(x)=\frac{1}{2}f(x;c_1,20000) + \frac{1}{2}f(x;b_1,20000)
\]
\[
c_1=(1,0,0,0); \quad b_1=(0,0,1,0)
\]
The density $p^5$:
\[
 p^{5}(x)=\sum_{i=0}^{9}\frac{1}{10}f(x;b_i,60)
\]
\[
b_j =
(\frac{1}{\sqrt{2}}\cos(\frac{2j\pi}{15}),
\frac{1}{\sqrt{2}}\sin(\frac{2j\pi}{15}),
\frac{1}{\sqrt{2}}\cos(\frac{j\pi}{15}),
\frac{1}{\sqrt{2}}\sin(\frac{j\pi}{15})), \text{ } j=0,...,9
\]

\subsection{Results for \ref{HyperbolicHeatKernel} and \ref{HeatSemigroupIdentity}}
\label{GenusGResults}
We present tables for KL averaged over 5 runs within one standard deviation along with testing of the heat semigroup property in this section. 

\subsubsection{KL Table}
\label{KLTableGenusG}
\begin{table}[H]
\centering
\caption{Comparisons of $\KL$ after 500 steps for hyperbolic heat kernel experiments}
\label{tab:KLcomparisonheatkernel}
\begin{tabular}{ccccc}
\toprule
t & x & Genus & Starting KL & Best KL \\
\midrule
0.5 & (0.0, 0.0) & 2 & 3.45 $\pm$ 1.52 & 0.03 $\pm$ 0.00 \\
0.5 & (0.0, 0.0) & 3 & 3.24 $\pm$ 0.92 & 0.03 $\pm$ 0.00 \\
0.5 & (0.0, 0.0) & 4 & 3.80 $\pm$ 0.52 & 0.05 $\pm$ 0.01 \\
0.05 & (0.0, 0.0) & 2 & 3.75 $\pm$ 0.74 & 0.06 $\pm$ 0.01 \\
0.05 & (0.0, 0.0) & 3 & 6.1 $\pm$ 1.59 & 0.11 $\pm$ 0.01 \\
0.05 & (0.0, 0.0) & 4 & 6.97 $\pm$ 0.84 & 0.08 $\pm$ 0.01 \\
1.5 & (0.0, 0.0) & 2 & 1.07 $\pm$ 0.21 & 0.08 $\pm$ 0.01 \\
1.5 & (0.0, 0.0) & 3 & 2.21 $\pm$ 0.24 & 0.12 $\pm$ 0.01 \\
1.5 & (0.0, 0.0) & 4 & 0.81 $\pm$ 0.12 & 0.11 $\pm$ 0.01 \\
0.05 & (0.5, 0.0) & 2 & 8.18 $\pm$ 1.24 & 0.15 $\pm$ 0.02 \\
0.05 & (0.5, 0.0) & 3 & 6.80 $\pm$ 0.74 & 0.16 $\pm$ 0.02 \\
0.05 & (0.5, 0.0) & 4 & 13.3 $\pm$ 1.26 & 0.06 $\pm$ 0.01 \\
0.05 & (0.5, 0.5) & 2 & 9.65 $\pm$ 1.18 & 0.64 $\pm$ 0.12 \\
0.05 & (0.5, 0.5) & 3 & 17.62 $\pm$ 1.24 & 0.30 $\pm$ 0.07 \\
0.05 & (0.5, 0.5) & 4 & 18.5 $\pm$ 0.89 & 0.18 $\pm$ 0.06\\
\bottomrule
\end{tabular}
\end{table}

\subsubsection{Relative Error of Hyperbolic Heat Kernel}
\label{RelErrorGenusG}
\begin{table}[H]
\centering
\caption{Testing heat semigroup property}
\label{tab:HeatSemigroup}
\begin{tabular}{ccccc}
\toprule
t & x & $x_0$ & Genus & Rel. Err. \\
\midrule
0.5 & (0.0, 0.0) & (0.0, 0.0) & 2 & 0.017 $\pm$ 0.003 \\
0.5 & (0.0, 0.0) & (0.5, 0.5) & 2 & 0.010 $\pm$ 0.004 \\
0.05 & (0.0, 0.0) & (0.0, 0.0) & 2 & 0.037 $\pm$ 0.012 \\
0.05 & (0.0, 0.0) & (0.5, 0.5) & 2 & 0.262 $\pm$ 0.025 \\
1.5 & (0.0, 0.0) & (0.0, 0.0) & 2 & 0.035 $\pm$ 0.007 \\
1.5 & (0.0, 0.0) & (0.5, 0.5) & 2 & 0.030 $\pm$ 0.007 \\
0.05 & (0.5, 0.0) & (0.5, 0.0) & 2 & 0.025 $\pm$ 0.006 \\
0.05 & (0.5, 0.0) & (0.0, 0.0) & 2 & 0.210 $\pm$ 0.014\\
0.05 & (0.5, 0.5) & (0.5, 0.5) & 2 & 0.056 $\pm$ 0.006 \\
0.05 & (0.5, 0.5) & (0.0, 0.0) & 2 & 0.069 $\pm$ 0.013 \\
0.05 & (0.5, 0.5) & (-0.5, 0.5) & 2 & 0.832 $\pm$ 0.056 \\
0.05 & (0.5, 0.5) & (-0.5, -0.5) & 2 & 0.23 $\pm$ 0.052 \\
0.05 & (0.5, 0.5) & (0.5, 0.5) & 4 & 0.086 $\pm$ 0.009 \\
0.05 & (0.5, 0.5) & (0.0, 0.0) & 4 & 0.667 $\pm$ 0.042 \\
0.05 & (0.5, 0.5) & (-0.5, 0.0) & 4 & 0.857 $\pm$ 0.073\\
\bottomrule
\end{tabular}
\end{table}
\begin{figure}
    \centering
    \includegraphics[width=0.4\textwidth]{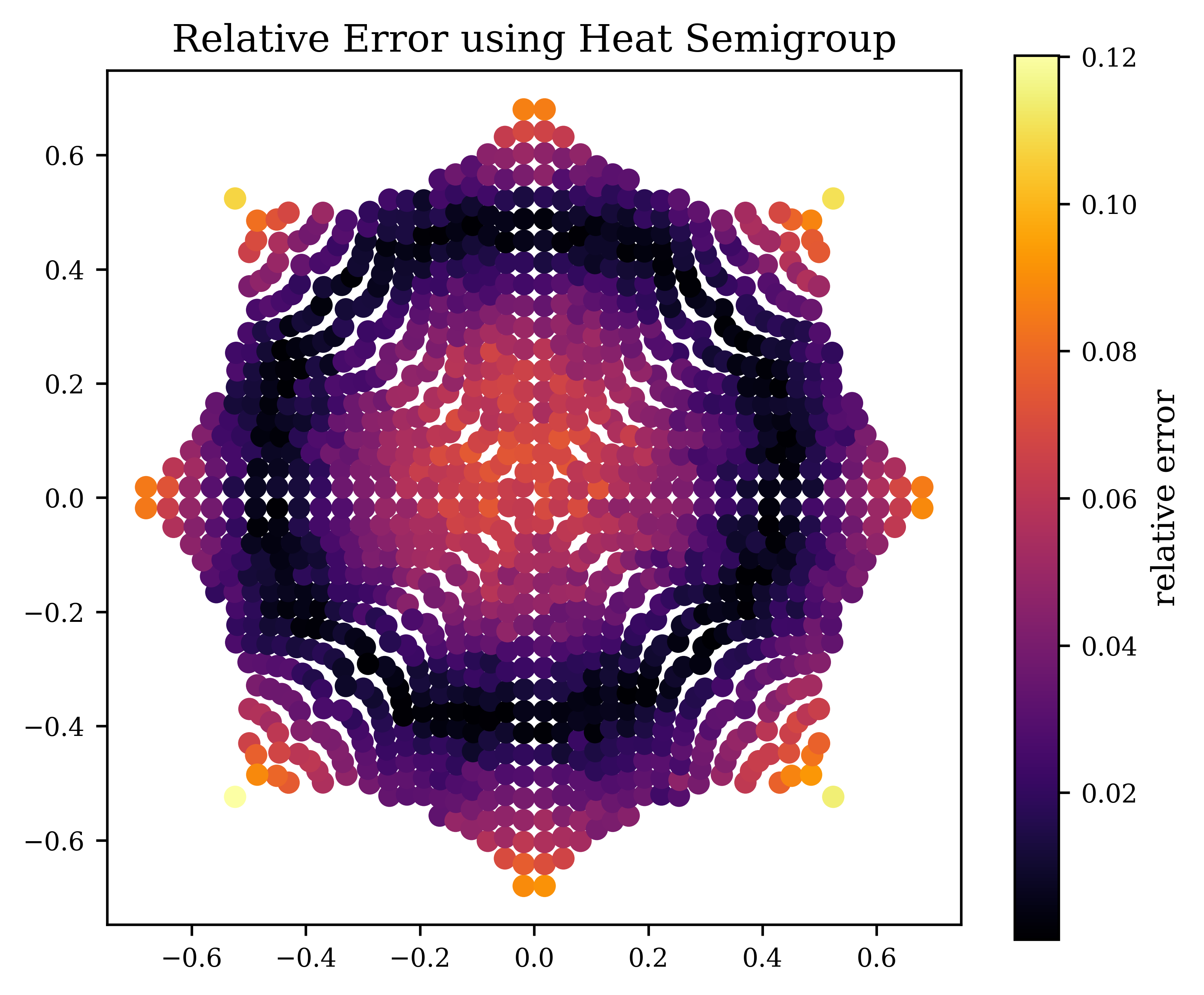}
    \includegraphics[width=0.4\textwidth]{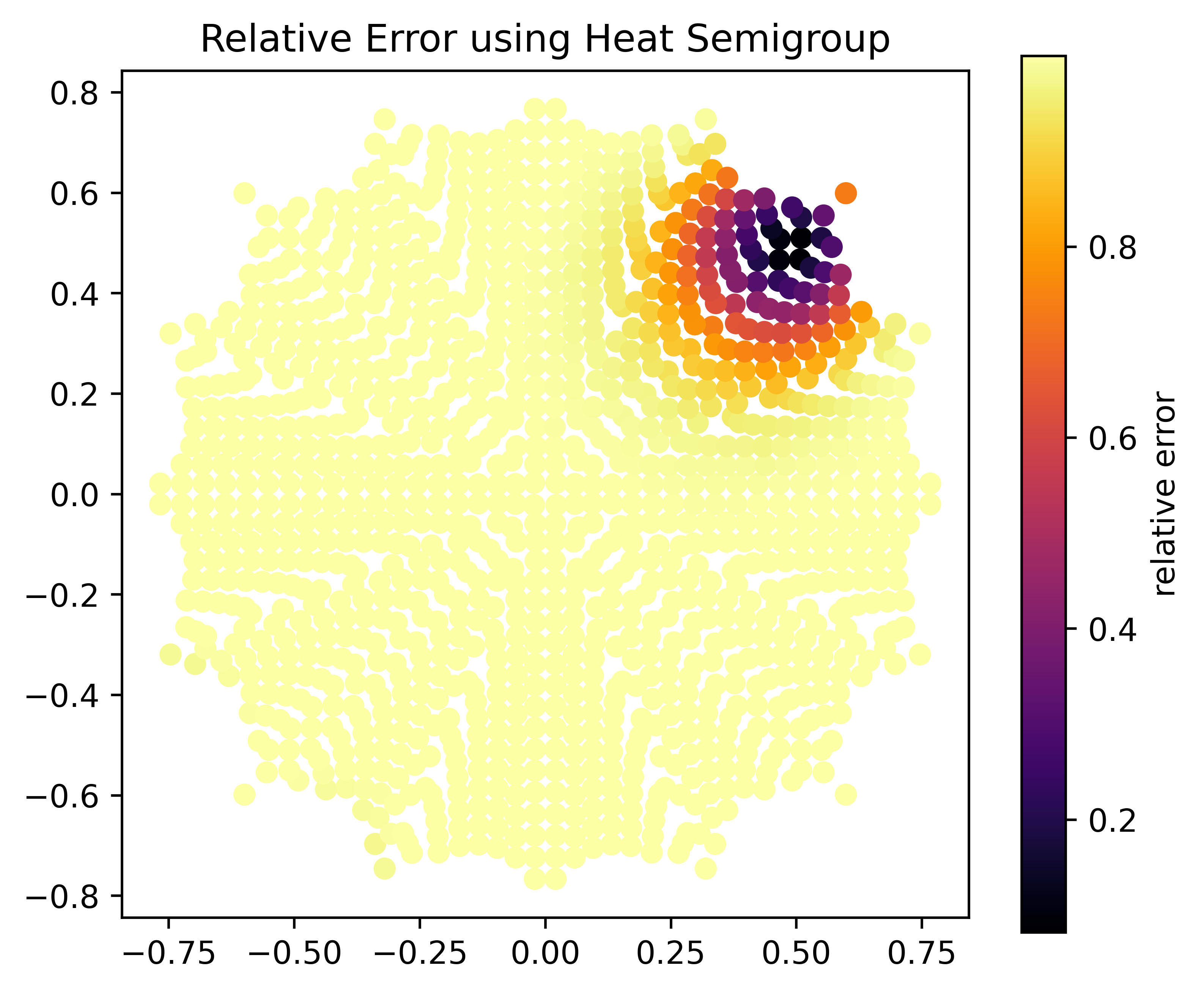}
    \caption{Relative Errors between MC Estimate using Heat Kernel with $t=0.5, x=(0.0,0.0)$ on $\Sigma_2$ (Left). Relative errors between MC estimate using hyperbolic heat kernel with $t=0.05, x=(0.5,0.5)$ on $\Sigma_4$ (Right).}
    \label{fig:heat_kernel_relErr}
\end{figure}
\[
\text{Rel. Err.} = \frac{\left|\frac{1}{N} \sum_{i=1}^N K_t^{\Sigma_g}(x_0,b_i) - K_{2t}^{\Sigma_g}(x_0, x) \right|}{K_{2t}^{\Sigma_g}(x_0, x)}, \quad b_i \sim B_{t,x}
\]
\subsection{KL results for Section \ref{SymmetrizedVMF}}
\label{KLLensSpace}
\begin{table}[H]
\centering
\caption{
Lens space KL and efficiency results for symmetrized vMF targets.
For each experiment, $p_{sym}^i$ denotes the $\mathbb Z_p$ symmetrization of the density $p^i$ defined in \ref{DensitiesLensSpaces}.
We report the final/best KL, the time required to reach KL $\leq 0.7$, and evaluation time. Results are averaged over 5 seeds within one standard deviation.
}
\label{tab:lens-space-sweep}
\begin{tabular}{ccllccc}
\toprule
Exp. & $(p,q)$ & Density & Model & Final KL & Time to KL $\leq 0.7$ (s) & Eval. time (s) \\
\midrule
1 & $(11,3)$ & $p^1_{sym}$ & PQ RealNVP & 0.05 $\pm$ 0.01 & 4 $\pm$ 2 & 9 $\pm$ 2 \\
  &          &                         & EMF     & 0.23 $\pm$ 0.08 & 493 $\pm$ 38 & 796 $\pm$ 64 \\
\midrule
2 & $(7,3)$ & $p^2_{sym}$ & PQ RealNVP & 0.28 $\pm$ 0.07 & 22 $\pm$ 4 & 90 $\pm$ 10 \\
  &         &                        & EMF        & 0.40 $\pm$ 0.12 & 1071 $\pm$ 82 & 1388 $\pm$ 121 \\
\midrule
3 & $(3,2)$ & $p^1_{sym}$ & PQ RealNVP & 0.06 $\pm$ 0.01 & 14 $\pm$ 3 & 72 $\pm$ 7\\
  &         &                        & EMF      & 0.73 $\pm$ 0.21 & not reached & 900 $\pm$ 88 \\
\midrule
4 & $(7,3)$ & $p^3_{sym}$ & PQ RealNVP & 0.09 $\pm$ 0.02 & 10 $\pm$ 3 & 48 $\pm$ 5 \\
  &          &                        & EMF    & 0.63 $\pm$ 0.19 & 641 $\pm$ 50 & 840 $\pm$ 84 \\
\bottomrule
\end{tabular}
\end{table}
\subsection{Learning Topology: a Case Study for $L(p;q)$}
\label{TopologyLearningL}
\begin{wrapfigure}{r}{0.5\textwidth}
  \centering
  \includegraphics[width=0.5\textwidth]{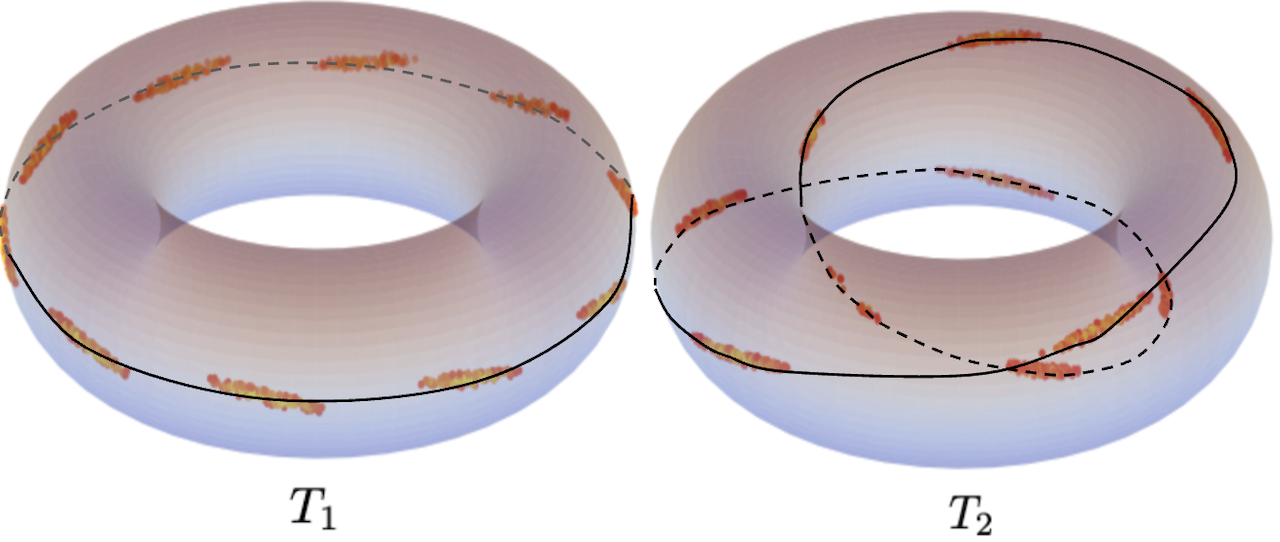}
  \caption{High density samples from the learned density approximating $p^5_{sym}$ on $L(3,2)$. The curve shows how boundary identification shears the longitude from $T_1$ to $T_2$; solid segments are front-facing and dotted segments are back-facing.}
  \label{fig:learningtopology}
\end{wrapfigure}
Recall that $L(p;q)$ is a boundary quotient of $T_1 \cup T_2$ via the map $A: \partial T_1 \rightarrow \partial T_2$. The densities we learn on $T_1 \cup_A T_2$ are naturally smooth on this boundary quotient. In Figure~\ref{fig:learningtopology}, we learn $p^5_{sym}$ with $(p,q)=(3,2)$ until the $\KL$ reached 0.07 (taking 208s, achieved using a scheduler). We observe that the $10$ modes that wrap around the boundary longitude, given by line in Figure~\ref{fig:learningtopology} under the solid torus $T_1$. The boundary map $A$ indicates that angular coordinate corresponding to $S^1$ in the product $S^1 \times D^2$ maps to $\theta \mapsto r \theta + s \phi$. In particular, in the topology is learned properly, we should see that the 10 modes on the boundary of $T_1$ are learned in a pattern in $T_2$ that wraps around the longitude $r=2$ and the meridian $s=1$ time. Following the curve in Figure~\ref{fig:learningtopology} on $T_2$, we see that this is exactly the case. Introducing the diagnostic
\[
\mathcal{E}_{\partial,\epsilon}=\mathbb{E}_{x\sim F_{*}Z_1}\left [\left |(1-I_2)q_1(\frac{x}{||x||+\epsilon})-I_2q_{2}(A\frac{x}{||x||+\epsilon})\right | \right ]
\]
where $\mu_{Z_1}$ is the prior on $Z_1$, $F_1$ is the learned diffeomorphism $F_1: Z_1 \rightarrow T_1$, $q_i$ is the learned density on $T_i$, and $I_2$ is $P(X\in T_2)$, $X$ being the target distribution. For the approximation of $p^5_{sym}$, we compute $\mathcal{E}_{\partial,0.0001}=0.04$. We obtain similar results for other approximations: for example, when $(p,q)=(7,3)$ with target density $p^3_{sym}$. Training to $\KL=0.03$, we have that $\mathcal{E}_{\partial,0.00001}=0.009$. While these numbers are attractive, these values depend heavily on $\epsilon$. The diagnostic demonstrates that topology is learned very close to the boundary, which is sufficient for visualizing the gluing map in Figure~\ref{fig:learningtopology} and thereby mapping high density boundary regions from one solid torus to another. That said, even for a good approximation in terms of $\KL$, the diagnostic demonstrates just as well that density values may roam freely as $\epsilon\rightarrow 0$. For example, for $p_{3}^{sym}$, if we use $\epsilon=0.000001$ instead of $0.00001$, $\mathcal{E}_{\partial,0.000001}=62065$. For $p^{5}_{sym}$, $\mathcal{E}_{\partial,0.00001}=43007$.

\subsection{Molecular p-fold symmetries and $L(2p;2p-1)$}
\label{Molecularp_foldsymmetries}
Identify $S^3$ with the unit quaternions $(w,x,y,z)=w+xi+yj+zk$. Identify $\mathbb{R}^3$ with $\text{Im}(\mathbb{H})=\{w+xi+yj+zk\in \mathbb{H}:w=0 \}$. Then $S^3$ acts on $\mathbb{R}^3$ by $h\cdot v=hvh^{-1}$ for $h\in S^3$, $v\in \mathbb{R}^3$ where multiplication in $hvh^{-1}$ is given by the group operation of $\mathbb{H}$. A molecular configuration in $\mathbb{R}^3$ may be specified by a vector $z_0\in \mathbb{R}^3$ together with an orthonormal basis $e_1,e_2$ for the plane defined by $z_0$. We write this configuration as $(z_0,e_1,e_2)$. Fixing an initial $(z_0, e_1, e_2)$, rotational molecular data may be given as points in $S^3$ - the data point $h\in S^3$ corresponds to the 3D molecular configuration corresponding to $h\cdot(z_0,e_1,e_2)=(h\cdot z_0,h\cdot e_1,h\cdot e_2)$ \cite{karney2007quaternions}. Suppose that $z_0=(1,0,0)=i$, $e_1=(0,1,0)=j$, and $e_2=(0,0,1)=k$. Let $\gamma=e^{2\pi i/2p}=\cos(\pi/p)+\sin(\pi/p)i\in S^3$. Acting by $\gamma$ on $\mathbb{R}^3$ corresponds to rotation by $2\pi/p$ about the normal vector $(1,0,0)=i$. Using the quaternion group action, we see that $\gamma\cdot (z_0,e_1,e_2)=(z_0,R_{2\pi/p}e_1,R_{2\pi/p}e_{2})$ where $R_{2\pi/p}$ is the $3\times 3$ matrix corresponding to rotation by $2\pi/p$ radians counterclockwise in the plane orthogonal to $z_0$. Write $I=(z_0,e_1,e_2)$. If a molecule exhibits energy/probability symmetries when rotated by $2\pi/p$ in the plane normal $z_0$, the (possibly unknown) density $m$ in molecular configuration coordinates should satisfy $m(h\cdot(\gamma \cdot I))=m(h\cdot I)$ for any $h\in S^3$. That is, $m((h\gamma) \cdot I)=m(h\cdot I)$. Viewing $m$ as defined on quaternions by removing the initial position $I$, we have that, for any $h\in S^3$, $m(h\gamma)=m(h)$. The action on $S^3$ generated by $h \mapsto h\gamma$ is precisely the $\mathbb{Z}_{2p}$ action for the normal covering space $\rho:S^3\rightarrow L(2p;2p-1)$ in standard $S^3$ coordinates.

\subsection{A Visualization on $\Sigma_4$}
\label{VisualizationSigma4}
\begin{figure}[H]
    \centering
    \includegraphics[width=\textwidth]{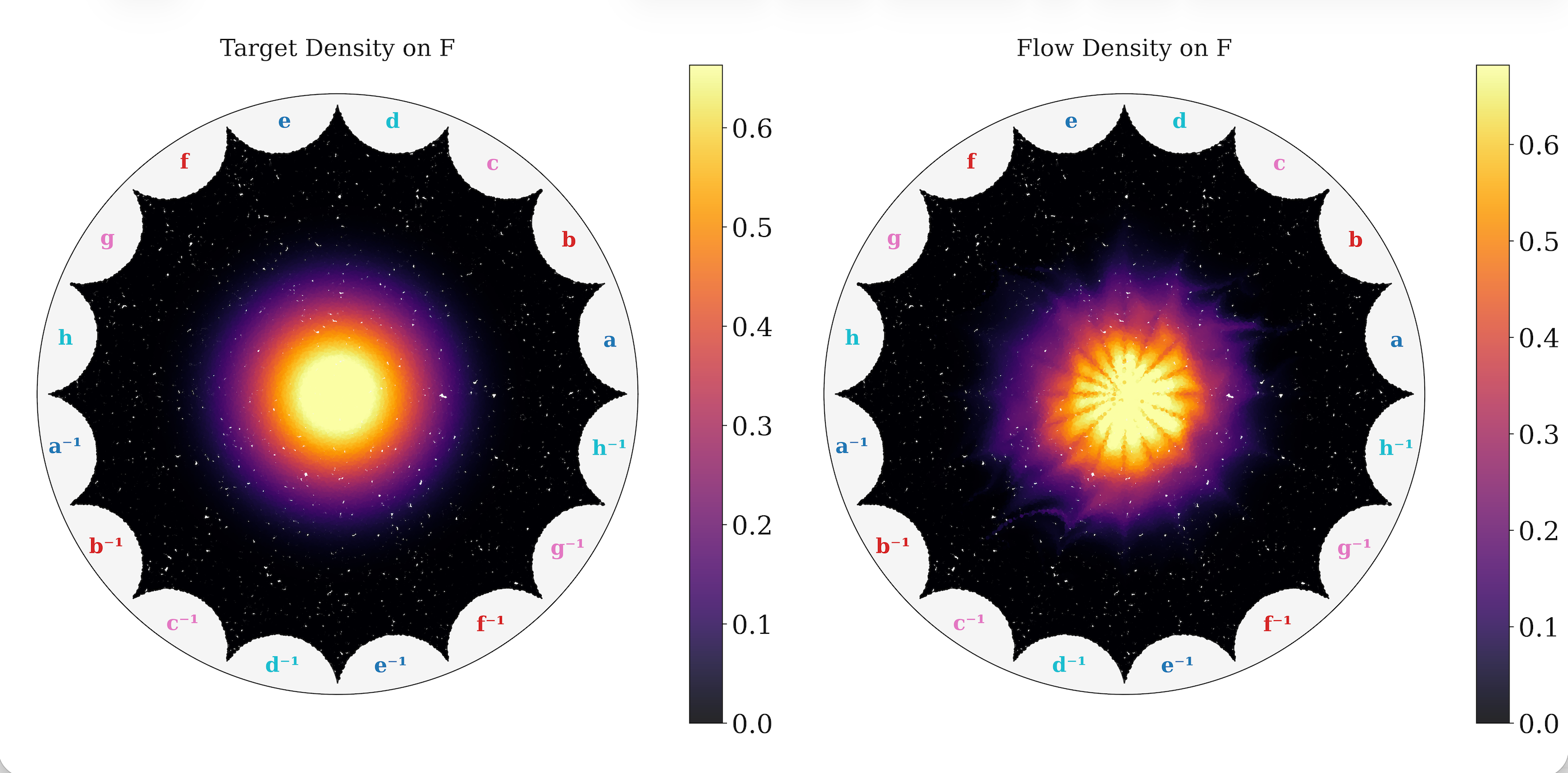}
    \caption{Hyperbolic heat kernel centered at $(0,0)$ with $t=0.1$ on $\Sigma_4$ showing $g > 2$ learning.}
    \label{fig:heatkernel16gon}
\end{figure}
\end{document}